\documentclass{article}

    \PassOptionsToPackage{sort, numbers, compress}{natbib} 

    \usepackage[final]{neurips_2024}

\usepackage[utf8]{inputenc} 
\usepackage[T1]{fontenc}    
\usepackage{hyperref}       
\usepackage{url}            
\usepackage{booktabs}       
\usepackage{amsfonts}       
\usepackage{nicefrac}       
\usepackage{microtype}      

\usepackage[usenames,dvipsnames,table,RGB]{xcolor}
\hypersetup{colorlinks}
\definecolor{BrickRed}{rgb}{0.6,0,0}
\definecolor{RoyalBlue}{rgb}{0,0,0.8}
\definecolor{Tdgreen}{rgb}{0,0.4,0.7}
\hypersetup{citecolor=MidnightBlue, linkcolor=BrickRed, 
  urlcolor=MidnightBlue}

\usepackage{amsmath}
\usepackage{graphicx} 
\usepackage{subcaption} 
\usepackage{algorithm}
\usepackage[noend]{algpseudocode}
\usepackage{wrapfig}
\usepackage{cleveref}
\usepackage{comment}
\usepackage{booktabs}
\usepackage{multirow}
\usepackage{makecell}

\newcommand{\system}{DEX}
\title{\system{}: Data Channel Extension for Efficient CNN Inference on Tiny AI Accelerators}

\makeatletter
\newcommand{\specificthanks}[1]{\@fnsymbol{#1}}

\newcommand{\nokia}{\textsuperscript{1}}
\newcommand{\unist}{\textsuperscript{2}}
\newcommand{\glasgow}{\textsuperscript{3}}

\author{%
Taesik Gong\nokia\unist\thanks{\textit{This work was done entirely while the author was affiliated with Nokia Bell Labs.}}  \quad Fahim Kawsar\nokia\glasgow \quad Chulhong Min\nokia
\vspace{1mm} \\
\nokia Nokia Bell Labs \quad \unist UNIST \quad \glasgow University of Glasgow
\vspace{1mm} \\
\texttt{taesik.gong@unist.ac.kr}\\
\texttt{\{fahim.kawsar, chulhong.min\}@nokia-bell-labs.com}
}

\begin{document}

\maketitle

\begin{abstract}


Tiny machine learning (TinyML) aims to run ML models on small devices and is increasingly favored for its enhanced privacy, reduced latency, and low cost. Recently, the advent of tiny AI accelerators has revolutionized the TinyML field by significantly enhancing hardware processing power. These accelerators, equipped with multiple parallel processors and dedicated per-processor memory instances, offer substantial performance improvements over traditional microcontroller units (MCUs). However, their limited data memory often necessitates downsampling input images, resulting in accuracy degradation. To address this challenge, we propose Data channel EXtension (\textit{\system{}}), a novel approach for efficient CNN execution on tiny AI accelerators. \system{} incorporates additional spatial information from original images into input images through patch-wise even sampling and channel-wise stacking, effectively extending data across input channels. By leveraging underutilized processors and data memory for channel extension, \system{} facilitates parallel execution without increasing inference latency. Our evaluation with four models and four datasets on tiny AI accelerators demonstrates that this simple idea improves accuracy on average by 3.5\%p while keeping the inference latency the same on the AI accelerator. The source code is available at \url{https://github.com/Nokia-Bell-Labs/data-channel-extension}.

\end{abstract}
\section{Introduction}\label{sec:intro}



Tiny machine learning (TinyML) is an active research field focused on developing and deploying machine learning models on extremely resource-constrained devices, such as microcontroller units (MCUs) and small IoT sensors. Compared to cloud-based AI, TinyML on devices offers benefits in privacy preservation, low latency, and low cost. 
While research efforts in TinyML, such as model compression techniques~\cite{han2015deep_compression, liu2017learning, liu2019metapruning, liberis2023differentiable, lin2017runtime, he2017channel}, have successfully reduced the size of AI models to fit into memory-constrained MCUs, the fundamental limitation in the processing capability of MCUs leads to long inference latency. This limitation hinders the widespread adoption of on-device AI, especially for real-time applications.

Recently, the advent of \textit{tiny AI accelerators} like the Analog Devices MAX78000~\cite{max78000} and Google Coral Micro~\cite{coralMicro} has revolutionized the TinyML field by dramatically boosting the model inference speed and leading a new phase of on-device AI. For instance, the MAX78000 AI accelerator~\cite{max78000} achieves 170$\times$ faster inference latency compared to an MCU processor (MAX32650~\cite{max32650}). 

To enable such acceleration, these tiny AI accelerators introduce several hardware optimization techniques. They often feature multiple convolutional processors (e.g., 64 processors in MAX78000~\cite{max78000}) and parallelize per-channel CNN operations across these processors. For further optimization, the memory architecture allows each processor to have a dedicated memory instance, i.e., \textit{per-processor memory instance}. This design enables simultaneous memory access to multiple channels from different processors. While these hardware-level optimizations bring significant performance improvements, we found that they also have several constraints at the expense of the optimizations. 
First, the per-processor memory architecture highly restricts the supported input image size because the data memory each processor can use for its input/output channels is limited to the capacity of its dedicated memory instance, which is a fraction of the total data memory divided by the number of processors. 
Consequently, most vision models for these accelerators are designed to support very small images, such as $32\times32$ pixels. Given that images captured by cameras are often generated with higher resolutions, downsampling is inevitable, leading to accuracy degradation due to information loss from the original image. Second, we found that processors and data memory are underutilized for the input layer due to the per-processor memory architecture; since input images typically have a low number of channels (e.g., RGB three channels), only a limited number of processors tied to memory instances are utilized while the remaining processors remain idle. For instance, on the MAX78000, 61 of 64 processors and per-processor memory instances remain unused in the first layer.

In this work, we propose a novel approach, Data channel EXtension (\emph{\system{}}), to overcome these constraints while still benefiting from the acceleration power of tiny AI accelerators. The core idea is to boost accuracy by extending the data channels to incorporate additional image information into unused data memory instances and processors, instead of simple downsampling. Owing to the parallel processing and memory access capabilities of tiny AI accelerators, our method can achieve this accuracy improvement without compromising inference latency. Specifically, \system{} involves two procedures: (1) pair-wise even sampling, where pixels from the original image are evenly sampled, and (2) channel-wise stacking, which arranges these samples across multiple channels.

To measure the impact of \system{} on accuracy and resource utilization, we conducted experiments on the MAX78000~\cite{max78000} and MAX78002~\cite{max78002} tiny AI accelerator platforms. \system{} was evaluated on four models, SimpleNet~\cite{hasanpour2016lets}, WideNet~\cite{hasanpour2016lets}, EfficientNetV2~\cite{tan2021efficientnetv2}, and MobileNetV2~\cite{sandler2018mobilenetv2}, using four vision datasets: ImageNette~\cite{imagenette}, Caltech101~\cite{caltech101}, Caltech256~\cite{caltech256}, and Food101~\cite{food101}. Our results show that \system{} improves average accuracy by 3.5\%p compared to the original model with downsampling and 3.6\%p compared to the existing coordinate augmentation approach (CoordConv~\cite{liu2018intriguing}), without increasing inference latency. Additionally, \system{} maximizes data memory and processor utilization, demonstrating its effectiveness in enhancing model performance on resource-constrained devices. In summary, \system{} can significantly enhance the performance of neural networks on tiny AI accelerators, leading to more efficient and effective deployment of AI on resource-constrained devices.


\section{Preliminary: tiny AI accelerators}\label{sec:background} 

\begin{figure}[t]
    \centering
    \begin{minipage}[t]{0.42\textwidth}
    \begin{subfigure}[t]{1\linewidth}
        \centering
        \includegraphics[width=\linewidth]{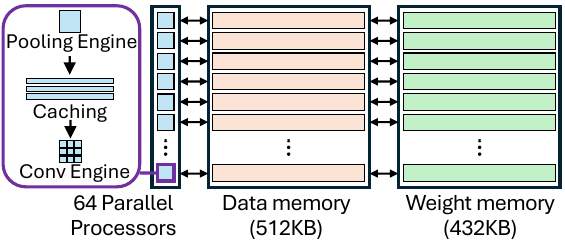}
    \end{subfigure}
    \caption{The architecture of a tiny AI accelerator (MAX78000~\cite{max78000}). 
    }\label{fig:acc_architecture}
    \end{minipage}
    \hspace{0.1cm}
    \begin{minipage}[t]{0.56\textwidth}
        \centering
        \includegraphics[width=\linewidth]{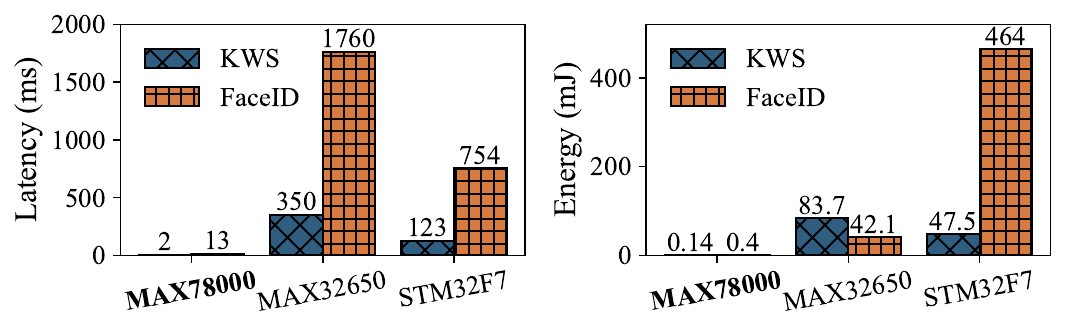}
        \caption{Comparison between an AI accelerator (MAX78000) and MCUs (MAX32650 and STM32F7).}\label{fig:mcu_performance}
    \end{minipage}
\end{figure}


The advent of tiny AI accelerators marks a pivotal shift towards on-device AI, greatly enhancing privacy and reducing latency. While a number of tiny-scale AI accelerators have emerged recently, such as Analog Devices MAX78000/MAX78002~\cite{max78000, max78002}, Google Coral Micro~\cite{coralMicro}, and GreenWaves GAP-8/GAP-9~\cite{gap89}, only a few are commercially available with access and control over their operations. In this paper, we focus on the MAX78000~\cite{max78000} and MAX78002~\cite{max78002} as our primary platforms since they are the most widely used tiny AI accelerator research platforms~\cite{moosmann2023tinyissimoyolo, ruegg2023kp2dtiny, gong2023collaborative, bakar2022protean, caronti2023fine, moss2022ultra} owing to the disclosed hardware details and open-source tools, enabling in-depth analysis and modification of their operations.



\paragraph{Architecture of tiny AI accelerators.} 

\begin{wrapfigure}{R}{0.4\textwidth}
    \centering
    \includegraphics[width=0.9\linewidth]{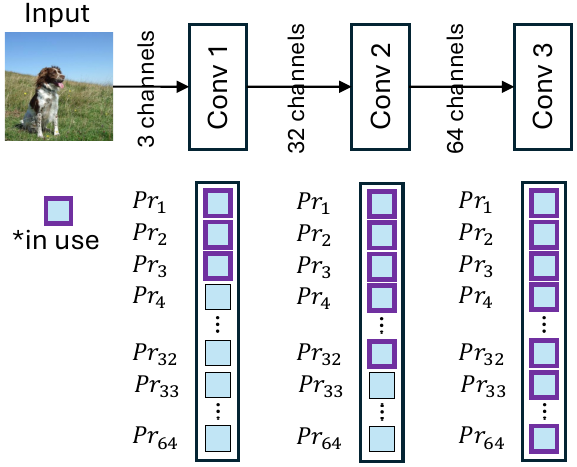}
    \caption{Processor utilization with varying input channels on the AI accelerator. 
    }
    \label{fig:acc_execution}
\end{wrapfigure}
The distinctive characteristic of tiny AI accelerators compared to conventional microcontroller units (MCUs) is \textit{parallel processors} that parallelize per-channel CNN operations across these processors. 
Figure~\ref{fig:acc_architecture} depicts an abstracted architecture of the MAX78000; MAX78002 has a similar architecture to MAX78000 with increased memory (1.3~MB data and 2~MB weight memory). Further details are in Appendix~\ref{app:exp_details:accelerator}. 
It has 64 parallel convolutional processors, each capable of performing specific operations independently. To maximize performance, each processor has a dedicated memory instance, i.e., \textit{per-processor memory instance} that optimizes data transfer with parallel access. For each CNN layer, operations on individual channels are assigned to separate convolutional processors and executed simultaneously, significantly reducing latency typically associated with convolutional algorithms. Each processor has a pooling engine, an input cache, and a convolution engine that can handle 3 by 3 kernels. The CNN accelerator includes 512 KB of data memory and 432 KB of weight storage memory. Within the 512 KB of data memory, an 8 KB per-processor memory instance is allocated to each of the 64 processors.
Figure~\ref{fig:acc_execution} shows the utilization of the processors ($Pr_i$) for executing CNNs with varying sizes of the input channels. Each processor communicates with a dedicated memory instance for each data channel. For example, given a three-channel image, three parallel processors are utilized in the first layer. 

\paragraph{Performance gain over MCUs.} 

A recent benchmark study~\cite{max78000benchmark} demonstrates the remarkable performance gain of the MAX78000 in terms of latency and energy consumption. Figure~\ref{fig:mcu_performance} shows that the MAX78000 significantly outperforms widely-used MCUs (MAX32650 with a Cortex-M4 at 120 MHz~\cite{max32650}, and a high-performance MCU, the STM32F7 with a Cortex-M7 at 216 MHz~\cite{stm32f7}) for face detection (FaceID) and keyword spotting (KWS). For KWS, latency is drastically reduced to only 2.0 ms, compared to 350 ms for the MAX32650 and 123 ms for the STM32F7. Accordingly, energy efficiency of the MAX78000 is also significant; it consumes only 0.40 mJ for FaceID, dramatically less than the 42.1 mJ and 464 mJ required by the MAX32650 and STM32F7, respectively.







\section{\system{}: Data channel extension for efficient CNN inference on AI accelerators}\label{sec:method}

\subsection{Constraints of per-processor memory instances in tiny AI accelerators for images}
As mentioned in~\S\ref{sec:background}, tiny AI accelerators leverage per-processor memory instances for faster data transfer with parallel access. However, we disclose that this causes several constraints at the expense of rapid data access: (1) low image resolution and (2) underutilized processors and data memory.



\paragraph{Low image resolution due to limited per-processor memory size.} 

MAX78000~\cite{max78000} has 512 KB data memory which is divided into 64 segments of 8 KB memory instances per processor, each storing the data of each input channel. This memory architecture highly restricts the supported input resolution.
For instance, an input image with a shape $3\times224\times224$ (channel, height, and weight), which is a typical size of ImageNet~\cite{imagenet}, does not fit the MAX78000 even with Q7 format (one byte for each value), as memory limit for each channel is 8 KB ($224\times224\sim50$ KB > 8 KB). Thus, the current practice on tiny AI accelerators is to shrink the resolution of input images by downsampling and accordingly, to design small models to process lower-resolution images, e.g., $3\times32\times32$. However, with this, it loses most of the information of the original image, which might lead to sub-optimal performance.

\begin{figure}[t]
    \centering
    \includegraphics[width=0.75\linewidth]{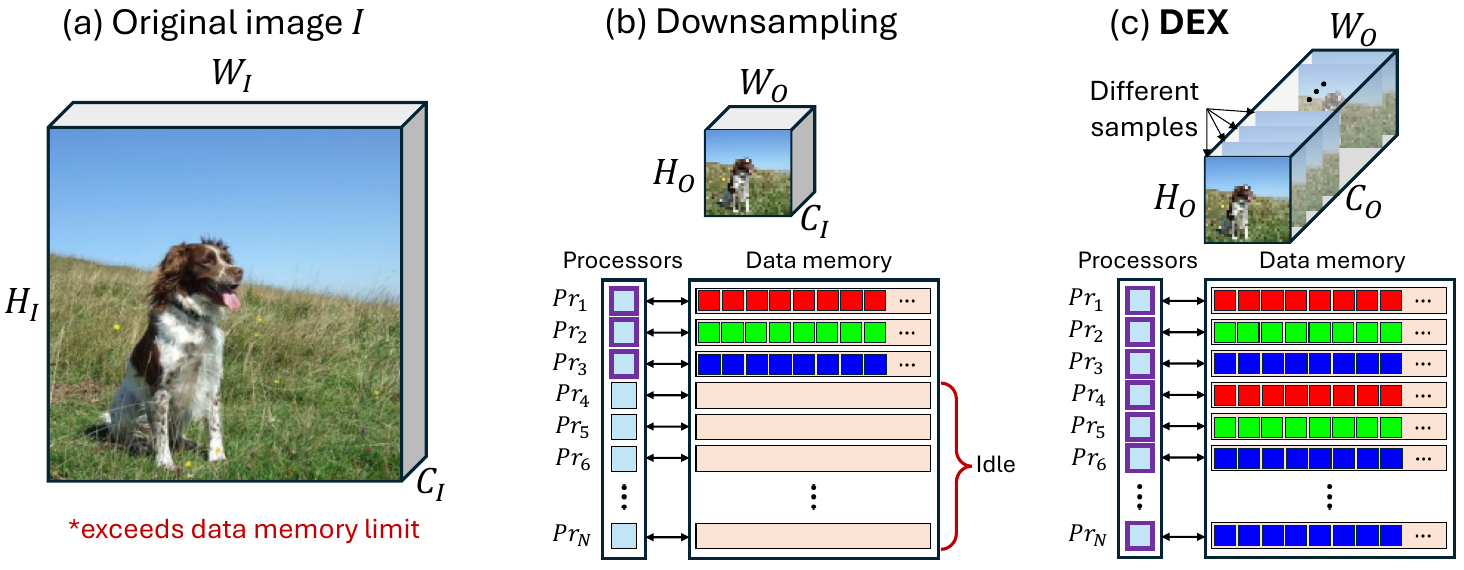}
    \caption{Comparison among different input data. (a) an original image that exceeds the data memory limit of the AI accelerator, (b) a downsampled image that fits the data memory but does not fully utilize parallel processors and data memory, and (c) a \system{}-generated image that incorporates more information from original image by extending data across channels with full utilization of parallel processors and data memory instances.}
    \label{fig:method_comparison}
\end{figure}

\paragraph{Underutilized processors and data memory for the input layer.}
Although per-processor memory instances allow simultaneous memory access from different processors, it also brings inefficiency in data memory and processor utilization, especially in the input layer.
Specifically, given an input image $I$ with the number of channels $C_I$, height $H_I$, and width $W_I$ (e.g., $3\times224\times224$) as shown in Figure~\ref{fig:method_comparison}(a), 
Figure~\ref{fig:method_comparison}(b) illustrates the downsampled image with the number of channels $C_I$, height $H_O$, and width $W_O$ (e.g., $3\times32\times32$), and its data memory usage in the AI accelerator. With three RGB channels, channel data are separately stored for each data memory instances for parallel execution. As there is $N$ processors and corresponding data memory instances, it leaves the remaining $N - 3$ processors and data memory instances idle. This provides an opportunity to utilize these idle data memory instances and parallel processors, which we detail in the following section.

\begin{figure}[t]
    \centering
    \includegraphics[width=0.9\linewidth]{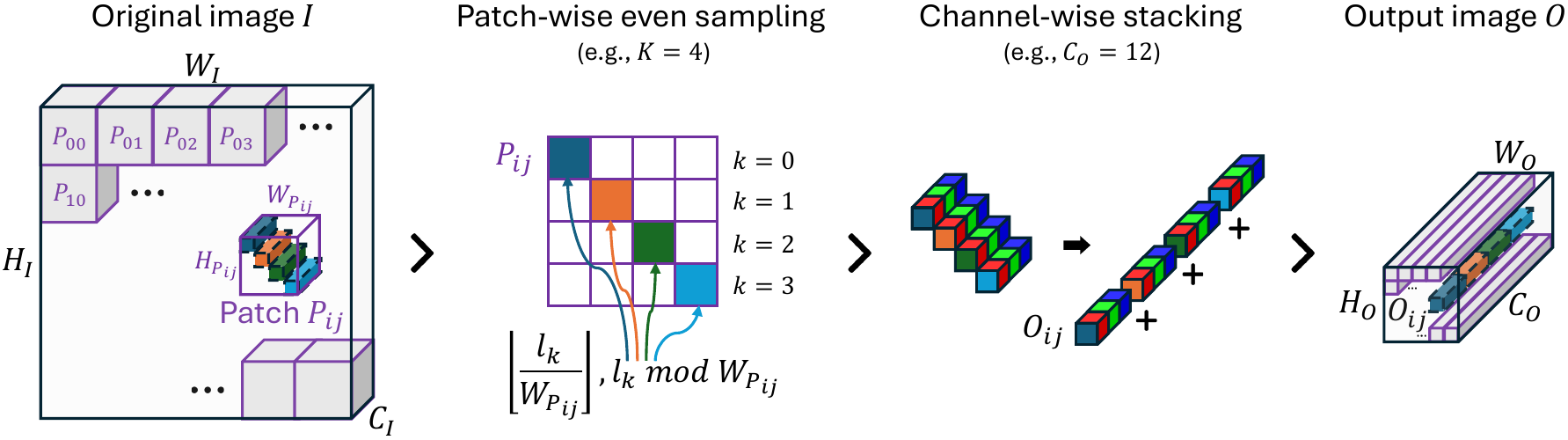}
    \caption{Overview of \system{}. \system{} divides the original image $I$ into multiple patches. \system{} then evenly samples pixels from each patch $P_{ij}$ and constructs an output pixel $O_{ij}$ by stacking samples across channels.}
    \label{fig:method}
\end{figure}

\subsection{\system{} procedure}
As aforementioned, we note two key observations: (1) the input image needs to be downsampled due to the limited memory of tiny AI accelerators, which means most of the pixel information cannot be utilized, and (2) there exist idle data memory instances and processors that could process up to $N$ channels in parallel. Although several recent studies have found efficient model architectures on tiny AI accelerators~\cite{moosmann2023tinyissimoyolo, ruegg2023kp2dtiny}, existing studies lack considerations on this inefficiency for input image processing in CNNs (further discussion on related work is in Appendix~\ref{sec:related_work}). 

Based on our observations, we propose Data channel EXtension (\textit{\system{}}) for efficient CNN execution on tiny AI accelerators. The key intuition behind \system{} is that we can utilize the remaining data memory to incorporate additional information from the original image into neural networks by extending the input data across channels. By utilizing this additional memory and processors, we can incorporate extra sample information for feature learning without sacrificing latency. Figure~\ref{fig:method_comparison}(c) shows the input data reshaped via \system{}, where each channel contains different pixel information from the original image. With \system{} extending data across channels (from $C_I$ to $C_O$), it can fully utilize the data memory and associated parallel processors. 
Figure~\ref{fig:method} shows an overview of the procedure of \system{}. Given an input image $I$ with a number of channels $C_I$, height $H_I$, and width $W_I$, \system{} generates an output image $O$ with an extended number of channels $C_O$, height $H_O$, and width $W_O$ (e.g., $64\times32\times32$) via patch-wise even sampling and channel-wise stacking. 



\paragraph{Patch-wise even sampling.}
The purpose of patch-wise even sampling is to select samples evenly spaced across the original image while keeping the spatial relationship among pixels. 
We first define a \textit{patch} from the original image in which a corresponding output pixel is generated. 
We denote i-th row and j-th column of patch $P_{ij}$ in $I$ as:
\begin{equation}
     P_{ij} = I \left[ \left\lfloor i \cdot \frac{H_I}{H_O} \right\rfloor : \left\lfloor (i+1) \cdot \frac{H_I}{H_O} \right\rfloor, \left\lfloor j \cdot \frac{W_I}{W_O} \right\rfloor : \left\lfloor (j+1) \cdot \frac{W_I}{W_O} \right\rfloor \right],
\end{equation}
where $[:, :]$ refers to a 2-D array slicing operation, specifying the selection of rows and columns sequentially. The number of patches is determined by the resolution of the output image, i.e., $H_O \times W_O$. For each patch $P_{ij}$, we generate the corresponding output data $O_{ij}$. This ensures that the spatial relationships among pixels in the input image are preserved in the output, maintaining spatial consistency throughout the process. 


The next step is to sample pixels within the patch considering the memory budget. Specifically, we define $K = \lceil \frac{C_O}{C_I} \rceil $ as the number of samples to be selected in each patch.
Given the height and width of patch $H_{P_{ij}} = \left\lfloor (i+1) \cdot \frac{H_I}{H_O} \right\rfloor - \left\lfloor i \cdot \frac{H_I}{H_O} \right\rfloor$ and $W_{P_{ij}} = \left\lfloor (i+1) \cdot \frac{W_I}{W_O} \right\rfloor - \left\lfloor i \cdot \frac{W_I}{W_O} \right\rfloor$, the i-th row and j-th column of output $O_{ij}$ can be represented by:
\begin{equation}
    O_{ij} = \left\{ P_{ij}\left[\left\lfloor \frac{l_k}{W_{P_{ij}}} \right\rfloor,\ \  l_k \mod W_{P_{ij}}\right] \mid l_k = k \cdot \left\lfloor \frac{H_{P_{ij}}\cdot W_{P_{ij}} - 1}{K - 1} \right\rfloor, \text{ for } k = 0, 1, \ldots, K-1 \right\},
\end{equation}
which means a collection of evenly distributed samples within each patch to encourage diverse information while minimizing the use of localized pixel information.
With patch-wise even sampling, selected samples are evenly distributed both across patches and within each patch.

\paragraph{Channel-wise stacking.} Channel-wise stacking arranges sampled data across multiple channels and keeps this procedure for all pixels to maintain data integrity. Channel-wise stacking is beneficial as it maintains consistency within each channel, preserving the spatial and contextual relationships of the sampled data. Specifically, after patch-wise even sampling, the samples are stacked across the channel axis in ascending order of the index $k$, and this is repeated for each $O_{ij}$. Note that $l_k = 0$ when $K = 1$, and this is identical to traditional downsampling. If $K > \frac{C_O}{C_I}$, it fills up the target channel with $P$'s data until the limit and discards the remaining channels. For instance, when using RGB channels $(C_I = 3)$ and if $C_O = 64$ and $K = 22$, it takes only the red channel for $i = 21$ and discards the remaining green and blue channels that exceed the channel limit of 64. Algorithm~\ref{alg:method} provides the pseudo-code that describes the procedure of \system{}'s channel extension algorithm.



\begin{algorithm}
\caption{\system{} Channel Extension Algorithm}\label{alg:method}
\begin{algorithmic}[1]
\Require{Source image $I$ in a shape $(C_I, H_I, W_I)$}
\Ensure{Reshaped image $O$  in a shape $(C_O, H_O, W_O)$}

\State $O \gets \text{zeros}(C_O, H_O, W_O)$

\For{$i \gets 0$ \textbf{to} $H_O-1$}
    \For{$j \gets 0$ \textbf{to} $W_O-1$}
        \State $\text{start\_row}, \text{end\_row} \gets \text{floor}(i \cdot \frac{H_I}{H_O}), \text{floor}((i + 1) \cdot \frac{H_I}{H_O})$
        \State $\text{start\_col}, \text{end\_col} \gets \text{floor}(j \cdot \frac{W_I}{W_O}), \text{floor}((j + 1) \cdot \frac{W_I}{W_O})$
        \State $P_{ij} \gets I[:, \text{start\_row}:\text{end\_row}, \text{start\_col}:\text{end\_col}]$ \Comment{Patch $P_{ij}$ of $I$}
        \State $K \gets \text{ceil}(\frac{C_O}{C_I})$ \Comment{Number of samples to be selected in $P_{ij}$}
        \For{$k \gets 0$ \textbf{to} $K-1$} \Comment{Get channels of $O_{ij}$ from $P_{ij}$}
            \State $H_{P_{ij}} \gets \text{end\_row}-\text{start\_row}$
            \State $W_{P_{ij}} \gets \text{end\_col}-\text{start\_col}$
            \State $l_k = k \cdot \text{floor}(\frac{H_{P_{ij}} \cdot W_{P_{ij}} -1}{K-1})$ 
            \State $ O[k\cdot C_I:(k+1)\cdot C_I,\ i,\ j] = P_{ij}[:, \text{floor}(\frac{l_k}{W_{P_{ij}}}), \ l_k \mod W_{P_{ij}}]$ 
        \EndFor
    \EndFor
\EndFor
\State \Return $O$
\end{algorithmic}
\end{algorithm}

\subsection{Further analysis on \system{}}\label{sec:method:analysis}

\begin{wrapfigure}{R}{0.55\textwidth}
    \centering
    \includegraphics[width=0.99\linewidth]{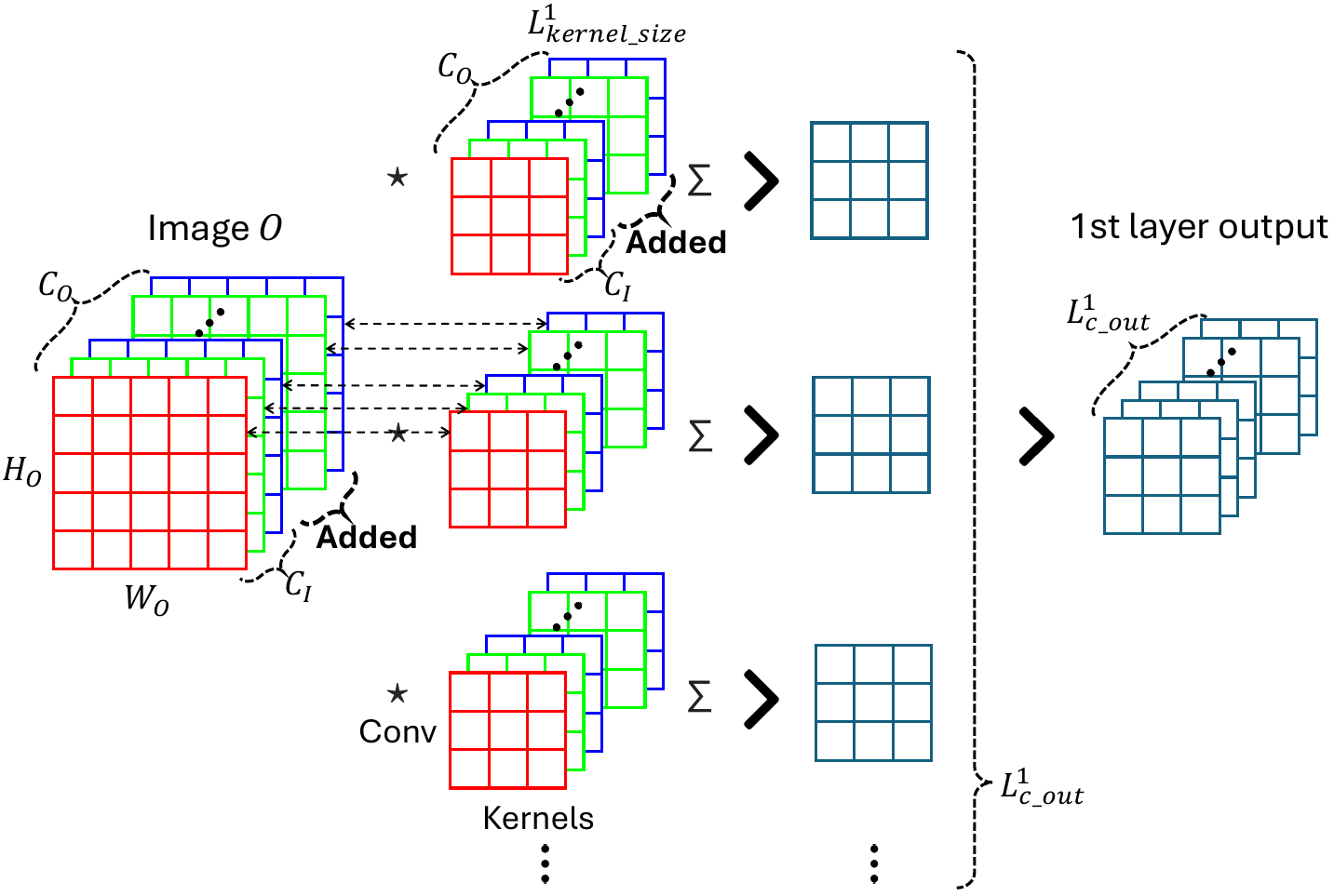}
    \caption{The initial CNN layer's operation with \system{}.}
    \label{fig:first_layer}
\end{wrapfigure}

\paragraph{Understanding how \system{} leads to performance improvement.}
\system{}'s ability to incorporate additional pixel information from the original image can improve the accuracy of CNNs. 
The extended channels provide further samples of adjacent areas in the original image, significantly broadening the receptive fields of features in the initial CNN layer. 
This expansion allows the model to detect more complex and subtle features early in the processing pipeline, which is critical for the nuanced understanding and interpretation of visual data. 
Specifically, Figure~\ref{fig:first_layer} visualizes how the first CNN layer operates with \system{}, where $L^1_{kernel\_size}$ and $L^1_{c\_out}$ refer to the kernel size and the output channel size of the first layer, respectively. 
It illustrates the application of the convolution operation across each enhanced channel ($C_O$ as opposed to $C_I$), where distinct kernel weights are applied to each channel. This ensures that the additional information is integrated into the output feature maps, thereby enriching the model's feature extraction capabilities. The convolutional layer processes the increased channel input, which is reflected in weight sums that construct output channels. 


\paragraph{Impact of channel extension on the number of parameters.}
Given the first CNN layer's kernel size $L^1_{kernel\_size}$ and the first layer's channel output size $L^1_{c\_out}$, the number of parameters required for the input layer can be calculated as $O_C \cdot L^1_{kernel\_size} \cdot L^1_{c\_out}$. If $O_C$ is 3, it is the same as the traditional downsampling without our channel extension. Note that this channel extension does not incur additional inference latency on the AI accelerator. We found that the channel extension increases $\sim3\%$ of the total parameters as we show in our experiment~\S\ref{sec:eval:result}. The rest of the layers remain the same. 
In addition to its simplicity, we have several reasons to change the first layer only, which we discuss further in \S\ref{sec:discussion}. 


\paragraph{Utilization of the original image information.}
With traditional downsampling, the utilization of the original input is $ \frac{H_O \cdot W_O}{H_I \cdot W_I} $, but with \system{}, this is extended to $ \frac{C_O}{C_I} \cdot \frac{H_O \cdot W_O}{H_I \cdot W_I} $. For instance, given a $3\times256\times256$ input image, a downsampled image $3\times32\times32$ utilizes only 1.6\% of the original information, while with \system{} and an output channel size $C_O = 64$, it can utilize 33.3\% of the original information. \system{} can accommodate all the information when $C_O = C_I\cdot\frac{H_I \cdot W_I}{H_O \cdot W_O}$

\paragraph{Maximum number of output channels.} Increasing the number of output channels allows \system{} to accommodate the original image information. The number of output channels denoted as $O_C$, that can be extended without increasing latency on AI accelerators is limited by the number of data memory instances $D_N$, i.e., $O_C < D_N$. For example, the MAX78000 has 64 data memory instances, allowing it to support up to $O_C = 64$ output channels without affecting inference latency.

\section{Evaluation}\label{sec:evaluation}

\subsection{Experimental settings}\label{sec:setting}

Here we explain experimental settings. Further details are in Appendix~\ref{app:exp_details}.

\paragraph{On-device testbed.} 
We evaluated \system{} on the off-the-shelf MAX78000 feather board~\cite{max78000fth} and MAX78002 Evaluation Kit~\cite{max78002evkit}, which are a development platform for the MAX78000~\cite{max78000} and MAX78002~\cite{max78002}, respectively, as shown in Figure~\ref{fig:testbed}. 
In this paper, we select these accelerators because they provide open-source tools for thorough analysis and modification of their internal processes, making them the most widely used tiny AI accelerator research platforms~\cite{moss2022ultra, moosmann2023tinyissimoyolo, ruegg2023kp2dtiny, gong2023collaborative, bakar2022protean, caronti2023fine}.


\paragraph{Model training and deployment.}
In our experiment, we use four models officially supported in the Analog Devices MAX78000/78002 Training framework~\cite{analogdevicesinc_ai8x_training}: SimpleNet~\cite{hasanpour2016lets}, WideNet~\cite{hasanpour2016lets}, EfficientNetV2~\cite{tan2021efficientnetv2}, and MobileNetV2~\cite{sandler2018mobilenetv2}.
The supported models from the framework were trained via quantization-aware training with 8-bit integers in PyTorch~\cite{paszke2019pytorch}. We follow the official training configuration (details in Appendix~\ref{app:exp_details:training}). The checkpoints are synthesized as embedded C codes for via the Analog Devices MAX78000/70002 Synthesis framework~\cite{analogdevicesinc_ai8x_synthesis}. SimpleNet and WideNet are developed for MAX78000 while EfficientNetV2 and MobileNetV2 are for MAX78002 considering the size of the models. All models are originally designed to take $3\times32\times32$ inputs, and \system{} increases the number of the channels in the first layer to 64. 

\paragraph{Datasets.}
We evaluated on four common vision datasets: (1) ImageNette~\cite{imagenette}, a ten-class subset of ImageNet~\cite{imagenet} with 9469/3925 train/test samples with the original image shape of $3\times350\times350$, (2) Caltech101~\cite{caltech101} with 101 objects classes having 6941/1736 train/test samples with the original image shape of $3\times300\times300$, (3) Caltech256~\cite{caltech256} with 256 objects classes having 23824/5956 train/test samples with the original image shape of $3\times300\times300$, and (4) Food101~\cite{food101} with 101 food categories with 75750/25250 train/test samples with the original image shape of $3\times512\times512$.

\paragraph{Baselines.} 
For baselines, we compare with the Downsampling method which is a straightforward way to reduce the size of the input under memory-constrained devices. It downsamples the input image to $3\times32\times32$. In addition, we compare \system{} with CoordConv~\cite{liu2018intriguing} which pointed out the limitation of traditional CNNs that relied on RGB images for the coordinate transformation problem and introduced the augmentation of $i$ and $j$ coordinates, which improved object detection efficiency by using two extra channels. The authors of CoordConv also introduced the third channel for an $r$ coordinate, where $r = \sqrt{(i - h/2)^2 + (j - w/2)^2}$, which they found effective in some experiments.


\subsection{Result}\label{sec:eval:result}

\begin{table}[t]
\caption{Average classification accuracy (\%) and corresponding standard deviations over three runs for each dataset and method. Bold type indicates those of the highest classification accuracy. }\label{tab:accuracy}
\centering
\resizebox{\columnwidth}{!}{%
\begin{tabular}{clccccc}
\toprule
\multicolumn{1}{c}{\textbf{Dataset}} & \multicolumn{1}{c}{\textbf{Method}} & \textbf{SimpleNet} & \textbf{WideNet} & \textbf{EfficientNetV2} & \textbf{MobileNetV2} & \textbf{AVG (\%)} \\
\midrule
\multirow{4}{*}{ImageNette} & Downsampling & 57.8 ± 1.2 & 61.8 ± 0.2 & 51.3 ± 0.5 & 62.0 ± 0.7 & 58.2 \\
 & CoordConv & 58.0 ± 1.1 & 61.7 ± 0.2 & 51.9 ± 0.1 & 61.6 ± 0.3 & 58.3 \\
 & CoordConv (r) & 55.4 ± 1.5 & 61.4 ± 0.2 & 51.7 ± 1.0 & 61.2 ± 1.1 & 57.4 \\
 & \textbf{DEX (ours)} & \textbf{61.4 ± 0.6} & \textbf{65.6 ± 0.6} & \textbf{56.8 ± 0.5} & \textbf{64.4 ± 0.6} & \textbf{62.0} \\
 \midrule
\multirow{4}{*}{Caltech101} & Downsampling & 54.6 ± 2.1 & 55.8 ± 1.2 & 38.6 ± 0.9 & 51.4 ± 1.6 & 50.1 \\
 & CoordConv & 53.8 ± 1.6 & 56.5 ± 0.1 & 38.7 ± 0.2 & 49.8 ± 0.5 & 49.7 \\
 & CoordConv (r) & 52.7 ± 0.5 & 56.0 ± 1.7 & 38.2 ± 1.0 & 49.7 ± 1.2 & 49.1 \\
 & \textbf{DEX (ours)} & \textbf{56.9 ± 1.3} & \textbf{61.1 ± 1.4} & \textbf{45.9 ± 1.9} & \textbf{53.3 ± 1.7} & \textbf{54.3} \\
 \midrule
\multirow{4}{*}{Caltech256} & Downsampling & 19.8 ± 0.6 & 20.8 ± 0.5 & 14.7 ± 0.4 & 22.4 ± 1.0 & 19.4 \\
 & CoordConv & 19.8 ± 0.5 & 21.3 ± 0.8 & 14.8 ± 0.8 & 22.7 ± 0.8 & 19.6 \\
 & CoordConv (r) & 20.0 ± 1.6 & 20.9 ± 0.6 & 14.5 ± 0.3 & 22.7 ± 0.4 & 19.5 \\
 & \textbf{DEX (ours)} & \textbf{22.8 ± 0.5} & \textbf{22.9 ± 0.9} & \textbf{18.3 ± 0.9} & \textbf{26.3 ± 0.5} & \textbf{22.6} \\
 \midrule
\multirow{4}{*}{Food101} & Downsampling & 16.0 ± 0.4 & 17.7 ± 0.7 & 12.1 ± 0.2 & 22.4 ± 0.6 & 17.1 \\
 & CoordConv & 16.1 ± 0.8 & 17.7 ± 0.3 & 12.0 ± 0.1 & 21.7 ± 0.3 & 16.9 \\
 & CoordConv (r) & 16.3 ± 0.4 & 17.3 ± 0.6 & 12.0 ± 0.6 & 20.9 ± 0.3 & 16.6 \\
 & \textbf{DEX (ours)} & \textbf{18.4 ± 0.4} & \textbf{20.9 ± 0.4} & \textbf{16.4 ± 0.1} & \textbf{23.3 ± 1.1} & \textbf{19.8}

 \\ \bottomrule
\end{tabular}%
}
\end{table}

\begin{table}[t]
\caption{Model size (Size), utilization of the original image information (InfoRatio), accelerator's processor utilization for the first layer (ProcUtil), and inference latency on the accelerator (Latency) for different models and methods averaged over three runs. }\label{tab:resource}
\centering
\resizebox{\columnwidth}{!}{%
\begin{tabular}{clrrrrr}
\toprule
\textbf{Model} & \multicolumn{1}{c}{\textbf{Method}} & \multicolumn{1}{c}{\textbf{InputChan}} & \multicolumn{1}{c}{\textbf{Size (KB)}} & \multicolumn{1}{c}{\textbf{InfoRatio ($\times$)}} & \multicolumn{1}{c}{\textbf{ProcUtil (\%)}} & \multicolumn{1}{c}{\textbf{Latency ($\mu s$)}} \\
\midrule
\multirow{4}{*}{SimpleNet} & Downsampling & 3 & 162.6 & 1.0 & 4.7 & 2592 ± 1 \\
 & CoordConv & 5 & 162.9 & 1.0 & 7.8 & 2592 ± 2 \\
 & CoordConv (r) & 6 & 163.0 & 1.0 & 9.4 & 2592 ± 2 \\
 & \textbf{DEX (ours)} & 64 & 171.2 & 21.3 & 100.0 & 2591 ± 1 \\
 \midrule
\multirow{4}{*}{WideNet} & Downsampling & 3 & 306.4 & 1.0 & 4.7 & 3820 ± 1 \\
 & CoordConv & 5 & 306.9 & 1.0 & 7.8 & 3820 ± 0 \\
 & CoordConv (r) & 6 & 307.1 & 1.0 & 9.4 & 3819 ± 1 \\
 & \textbf{DEX (ours)} & 64 & 319.3 & 21.3 & 100.0 & 3818 ± 1 \\
 \midrule
\multirow{4}{*}{EfficientNetV2} & Downsampling & 3 & 742.4 & 1.0 & 4.7 & 11688 ± 2 \\
 & CoordConv & 5 & 743.0 & 1.0 & 7.8 & 11685 ± 3 \\
 & CoordConv (r) & 6 & 743.2 & 1.0 & 9.4 & 11689 ± 1 \\
 & \textbf{DEX (ours)} & 64 & 759.6 & 21.3 & 100.0 & 11690 ± 2 \\
 \midrule
\multirow{4}{*}{MobileNetV2} & Downsampling & 3 & 1317.8 & 1.0 & 4.7 & 3553 ± 4 \\
 & CoordConv & 5 & 1318.2 & 1.0 & 7.8 & 3554 ± 1 \\
 & CoordConv (r) & 6 & 1318.4 & 1.0 & 9.4 & 3554 ± 2 \\
 & \textbf{DEX (ours)} & 64 & 1330.7 & 21.3 & 100.0 & 3552 ± 3
 \\ \bottomrule
\end{tabular}%
}
\end{table}

\paragraph{Overall accuracy.}

Table~\ref{tab:accuracy} shows the overall accuracy for four different datasets with the baselines and \system{}. As shown, extending data channels to utilize additional input information improves accuracy in \system{}. Specifically, \system{} achieved 3.5\%p higher accuracy compared to downsampling and 3.6\% higher accuracy compared to CoordConv across datasets. CoordConv shows lower accuracy compared with downsampling (0.1\%p degradation on average), showing they are not very effective solutions. This finding aligns with previous results indicating that CoordConv is useful for specific tasks such as object detection, where coordinate information is important~\cite{liu2018intriguing}. We found CoordConv (r) has a similar pattern to CoordConv. Overall, \system{}'s accuracy improvement shows the effectiveness of using extra information from the original image for feature learning.


\paragraph{Resource usage.} 
Table~\ref{tab:resource} compares the resource usage of the baseline and \system{}. First, we found that, although \system{} extends the number of channels in the first CNN layer to 64, its impact on the model size is negligible (an average increment of 3.2\% compared to no channel extension). \system{} utilizes 21.3$\times$ more image information compared to downsampling, which is the primary reason for the accuracy improvement. As expected, \system{} does not increase on-device inference latency, even though it maximally utilizes the processors on the AI accelerators for information processing. 
This result is consistent across the four datasets, as all the models are designed to take the same input size in MAX78000 and MAX78002. 

\begin{figure}[t]
    \centering
    \includegraphics[width=1\linewidth]{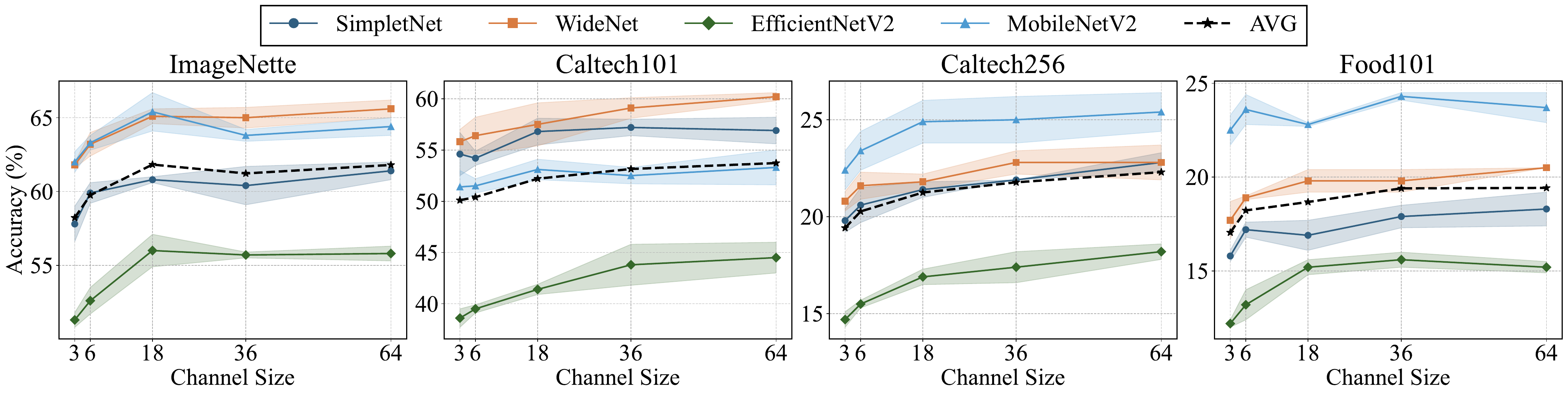}
    \caption{Accuracy of \system{} varying the channel size. The shaded areas are standard deviations.}
    \label{fig:accuracy_channel}
\end{figure}

\begin{table}[t]
\centering
\caption{Model size (Size) with relative increment (\%) compared to the three channels and average inference latency on the accelerator (Latency) with standard deviations over three runs, varying the channel size.}\label{tab:resource_channel}
\resizebox{\columnwidth}{!}{%
\begin{tabular}{lcrrrrr}
\toprule
 & \textbf{Model} & \multicolumn{1}{c}{\textbf{Chan = 3}} & \multicolumn{1}{c}{\textbf{Chan = 6}} & \multicolumn{1}{c}{\textbf{Chan = 18}} & \multicolumn{1}{c}{\textbf{Chan = 36}} & \multicolumn{1}{c}{\textbf{Chan = 64}} \\
 \midrule
& SimpleNet & 162.6 & 163.0 \small(+0.3\%) & 164.7 \small(+1.3\%) & 167.3 \small(+2.9\%) & 171.2 \small(+5.3\%) \\
& WideNet & 306.4 & 307.1 \small(+0.2\%) & 309.6 \small(+1.0\%) & 313.4 \small(+2.3\%) & 319.3 \small(+4.2\%) \\
& EfficientNetV2 & 742.4 & 743.2 \small(+0.1\%) & 746.6 \small(+0.6\%) & 751.7 \small(+1.3\%) & 759.6 \small(+2.3\%) \\
\multirow{-4}{*}{\textbf{Size \small(KB)}} & MobileNetV2 & 1317.8 & 1318.4 \small(+0.0\%) & 1321.0 \small(+0.2\%) & 1324.8 \small(+0.5\%) & 1330.7 \small(+1.0\%) \\
 \midrule
\multicolumn{1}{c}{} & SimpleNet & 2592 ± 1 & 2592 ± 2 & 2591 ± 1 & 2590 ± 1 & 2591 ± 1 \\
\multicolumn{1}{c}{} & WideNet & 3820 ± 1 & 3820 ± 2 & 3825 ± 1 & 3819 ± 3 & 3818 ± 1 \\
\multicolumn{1}{c}{} & EfficientNetV2 & 11688 ± 2 & 11691 ± 2 & 11692 ± 3 & 11691 ± 0 & 11690 ± 2 \\
\multicolumn{1}{c}{\multirow{-4}{*}{\textbf{Latency ($\mu s$)}}} & MobileNetV2 & 3553 ± 4 & 3553 ± 1 & 3552 ± 1 & 3554 ± 0 & 3552 ± 3
\\\bottomrule
\end{tabular}%
}
\end{table}

\paragraph{Accuracy according to the channel size.} We varied the size of the channels from 3 (downsampling) to 6, 18, 36, and 64 with \system{} to understand the impact of the channel size in terms of accuracy. Figure~\ref{fig:accuracy_channel} shows the accuracy variation according to the channel size across the four datasets. As shown, it seems that a higher number of channels increases accuracy in general. This means that selecting the highest channel size supported in AI accelerators might be an effective strategy in practice, considering that it does not incur the latency increase. Still, there are some cases where the accuracy of the highest channel size (64) is not the best among them. This means there might be an optimal number of channels tailored to a specific dataset and model architecture, which might be found in the model development process. 

\paragraph{Resource usage according to the channel size.}

\begin{wrapfigure}{o}{0.55\textwidth}
    \centering
    \begin{subfigure}[b]{0.27\textwidth}
        \centering
        \includegraphics[width=\textwidth]{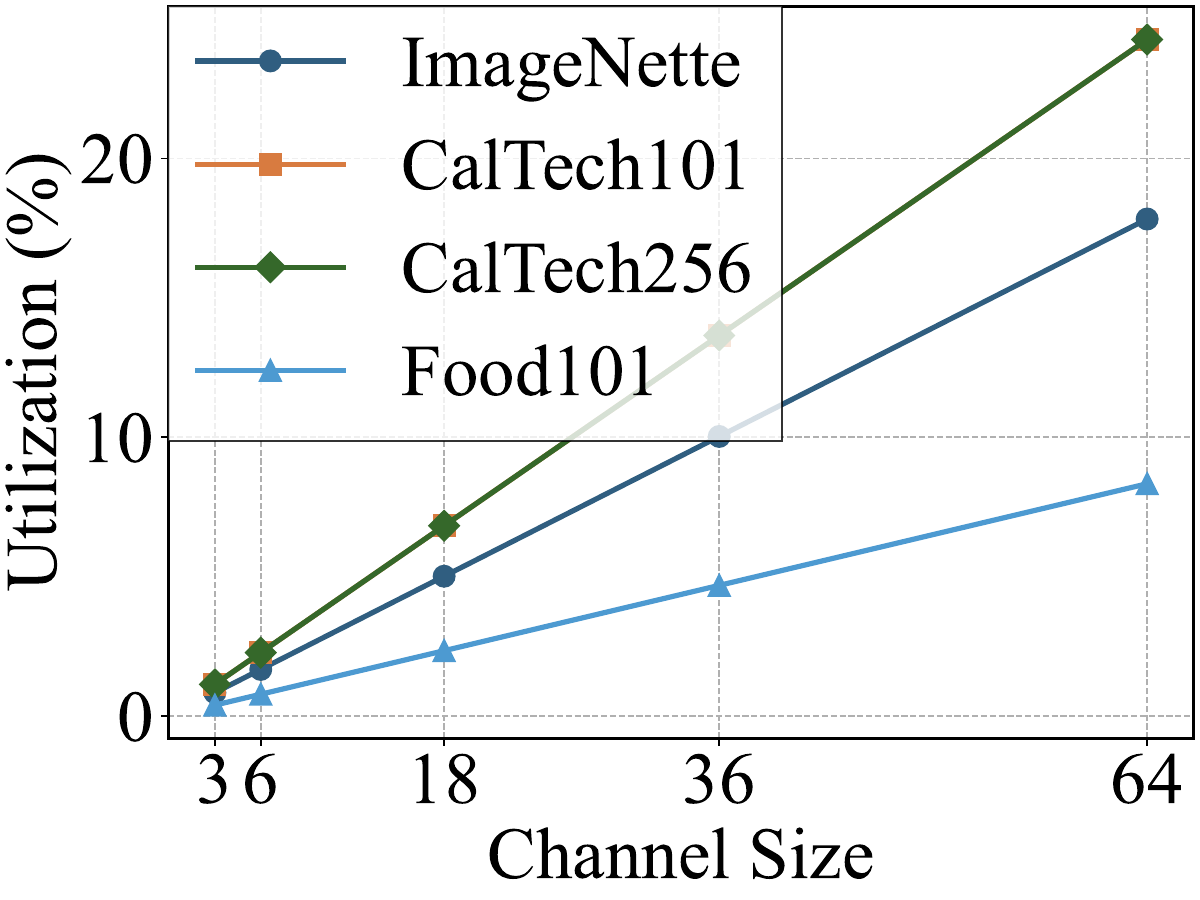} 
        \caption{Information utilization.}
        \label{fig:resource_channel:info}
    \end{subfigure}
    \hfill 
    \begin{subfigure}[b]{0.27\textwidth}
        \centering
        \includegraphics[width=\textwidth]{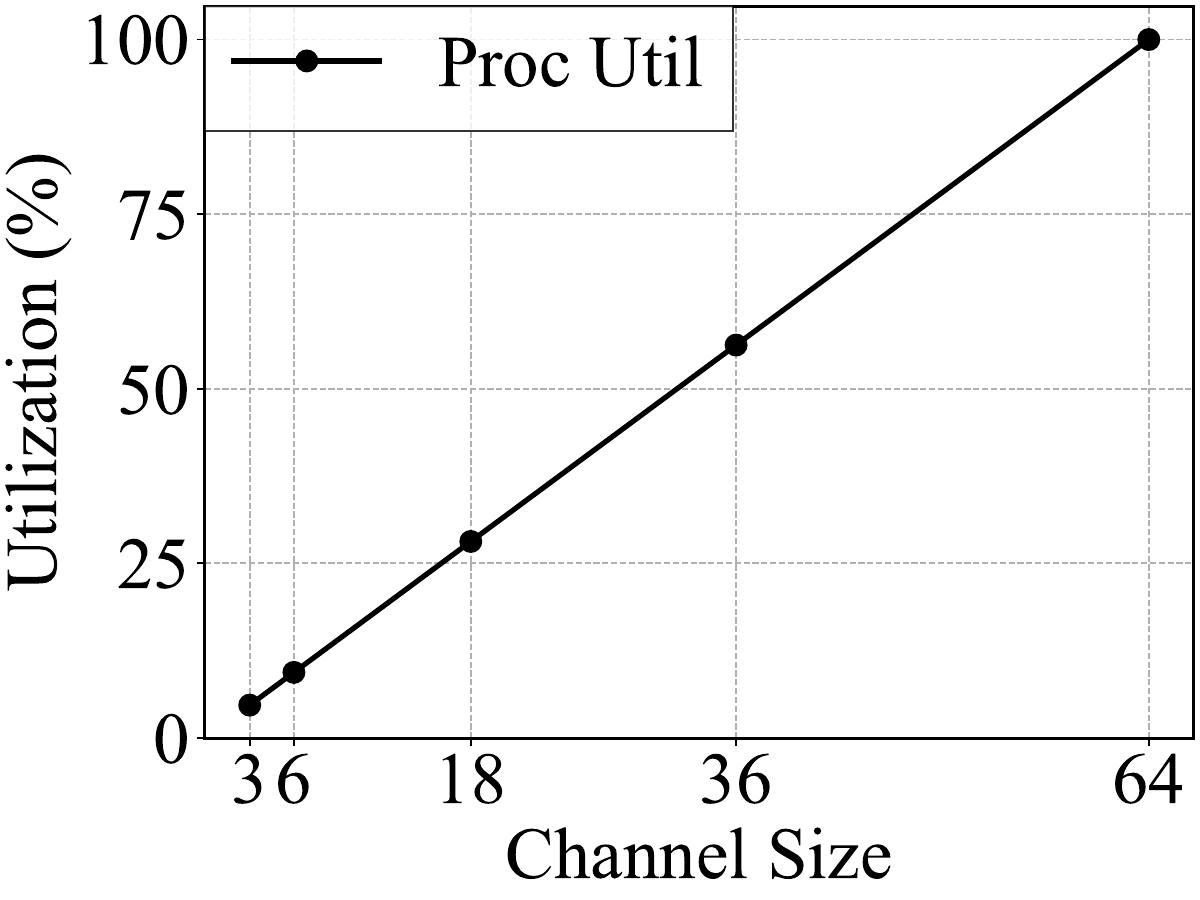} 
        \caption{Processor utilization.}
        \label{fig:resource_channel:proc}
    \end{subfigure}
    \caption{Resource usage varying the channel size.}\label{fig:resource_channel}
\end{wrapfigure}

We also measure resource usage varying the channel size. 
First, we measured the model size and inference latency as shown in Table~\ref{tab:resource_channel}. The model size increment is negligible and inference latency remains the same across different numbers of channels. The model size and inference latency are the same for the four datasets as all the models are designed to take the same input size in MAX78000 and MAX78002. Second, we measure the information utilization from the original image and processor utilization in the AI accelerators (Figure~\ref{fig:resource_channel}). 
The utilization of the original image information depends on the size of the original data size, which grows linearly according to the channel size. We found a correlation between information utilization rate and accuracy improvement. For example, Caltech101 and Caltech256 had utilization rates of 24.3\%, improving accuracy by 4.2\%p and 3.2\%p, respectively, while Food101 had an 8.3\% utilization rate with a 2.7\%p accuracy improvement.
The processor utilization linearly increases until 100\% with 64 channels size, which is the number of parallel processing units in the evaluated platforms.

\begin{wraptable}{r}{0.5\textwidth}
    \vspace{-0.4cm}
\caption{Comparison of data extension strategies.}\label{tab:ablation}
\centering
\resizebox{\linewidth}{!}{%
\begin{tabular}{lrrc}
\toprule

\multicolumn{1}{c}{\textbf{Method}} & \multicolumn{1}{c}{\textbf{InputChan}} & \multicolumn{1}{c}{\textbf{InfoRatio ($\times$)}} & \textbf{Accuracy} \\
\midrule
Downsampling & 3 & 1.0 & 57.8 ± 1.2 \\
Repetition & 64 & 1.0 & 56.3 ± 0.8 \\
Rotation & 64 & 1.0 & 55.4 ± 0.7 \\
Tile per channel & 64 & 21.3 & 39.4 ± 0.7 \\
Patch-wise seq. & 64 & 21.3 & 61.0 ± 1.5 \\
Patch-wise rand. & 64 & 21.3 & 60.4 ± 1.0 \\
\textbf{DEX} & 64 & 21.3 & \textbf{61.4 ± 0.6}
\\ \bottomrule
\end{tabular}%
}
\end{wraptable}
\paragraph{Comparison of alternative data extension strategies in \system{}.}  To understand the effectiveness of our patch-wise even sampling and channel-wise stacking, we compared \system{} with other possible data channel extension strategies. We compared with four strategies: repeating the same downsampled image across the channels (Repetition), generating slightly different images through rotation (Rotation), dividing the original image into multiple tiles and stacking those tiles across channels (Tile), patch-wise sequential sampling (Patch-wise seq.) that samples pixels sequentially within a patch, and patch-wise random sampling (Patch-wise rand.) that randomly samples within a patch. Further implementation details are in Appendix~\ref{app:alternative:details}. In this experiment, we used SimpleNet and evaluated it on ImageNette. Table~\ref{tab:ablation} shows the results. Repetition does not improve accuracy over downsampling, indicating that merely increasing the number of kernels does not lead to performance gains. Rotation shows a slight decrease in accuracy compared to Repetition, which suggests that slight changes through rotation do not enhance performance. Interestingly, Tile shows low accuracy, demonstrating the importance of having a complete view of the original image in each channel, rather than focusing on specific regions. Both Patch-wise sequential and Patch-wise random samplings show lower accuracy than \system{}'s patch-wise even sampling, highlighting the importance of even sampling for better performance.



\section{Related work}\label{sec:related_work}

\paragraph{TinyML.}
Tiny Machine Learning (TinyML) is an emerging field that focuses on adapting machine learning techniques for highly resource-constrained devices, such as microcontroller units (MCUs). These devices often come with limited memory, typically hundreds of kilobytes of SRAM. Research in this area has mostly concentrated on reducing model size through various compression techniques, such as model pruning~\cite{liberis2023differentiable, han2015deep_compression, lin2017runtime, he2017channel, liu2017learning, liu2019metapruning}, model quantization~\cite{han2015deep_compression, rusci2020memory, choi2018pact, wang2019haq, rastegari2016xnor, zhou2016dorefa, zhu2016trained}, and neural architecture search (NAS)~\cite{liberis2021munas, fedorov2019sparse, cai2018proxylessnas, cai2019once}.
In addition, several studies have explored the efficient utilization of memory resources (e.g., SRAM). Examples include optimizing on-device training processes~\cite{kwon2023tinytrain, lin2022device} and designing memory-efficient neural architectures~\cite{lin2021mcunetv2, zheng2024streamnet}. Unlike these approaches that primarily target MCUs, our research utilizes the distinctive architecture of tiny AI accelerators to enhance both memory efficiency and overall performance.



\paragraph{Tiny AI accelerators.}

Several studies have leveraged tiny AI accelerators for small-scale on-device AI applications. For instance, TinyissimoYOLO~\cite{moosmann2023tinyissimoyolo} offers a quantized, memory-efficient, and ultra-lightweight object detection network, showcasing its effectiveness on the MAX78000 platform. Additionally, KP2DTiny~\cite{ruegg2023kp2dtiny} introduces a quantized neural keypoint detector and descriptor specifically optimized for MAX78000 and Coral AI accelerators. Moreover, Synergy represents a multi-device collaborative inference platform across wearables equipped with tiny AI accelerators~\cite{gong2023collaborative}.
Another line of studies utilized tiny AI accelerators in battery-free or intermittent computing scenarios~\cite{bakar2022protean, caronti2023fine}. Traditionally, hardware accelerators on low-power AI platforms were capable of only one-shot atomic executions of a neural network inference without intermediate result backups. A study proposed a toolchain to address this that allows neural networks to execute intermittently on the MAX78000 platform~\cite{caronti2023fine}.
To the best of our knowledge, there has been no work that manipulates data and models to efficiently utilize computing resources considering the unique architecture of tiny AI accelerators.


\paragraph{Image channel extension in CNNs.}
Several studies have explored augmenting images with additional information to construct multi-channel inputs for Convolutional Neural Networks (CNNs). Liu et al. proposed a multi-modality image fusion approach, combining visible, mid-wave infrared, and motion images for enhanced object detection~\cite{liu2017multi}, while Wang et al. presented depth-aware CNN for image segmentation~\cite{wang2018depth}. These approaches require extra sensing channels to acquire data, such as infrared cameras and depth cameras. 
Similarly, other research has incorporated location data to improve performance for segmentation~\cite{wang2018location} and object detection tasks~\cite{liu2018intriguing}. For instance, CoordConv~\cite{liu2018intriguing} pointed out the limitation of traditional CNNs that relied solely on RGB images for the coordinate transformation problem and introduced the augmentation of $i$ and $j$ coordinates, which improved object detection efficiency. However, these methodologies often necessitate additional sensor modalities or are tailored for specific applications such as object detection, which restricts their general use. Nevertheless, adapting findings from those studies within \system{} could be an interesting future direction.



\section{Discussion and conclusion}\label{sec:discussion}

We introduced \system{}, a novel method to enhance CNN efficiency on tiny AI accelerators by augmenting input data across unused memory. Evaluations on four image datasets and models showed that \system{} improves accuracy without increasing inference latency. This method maximizes the processing and memory capabilities of tiny AI accelerators, making it a promising solution for efficient AI model execution on resource-constrained devices.

\paragraph{Limitations and potential societal impacts.}
We modified only the initial CNN layer due to simplicity, effectiveness, and memory constraints. The first layer, representing image data in three channels (RGB), has the most unused processors after initial data assignment. Extending channels at the first layer significantly increases data utilization with minimal impact on model size. This approach aligns with the design of weight memory in tiny AI accelerators, which maximizes model capacity by collective use across processors. We think \system{} might be less effective in certain tasks where incorporating more pixel information is not beneficial. In those cases, alternative data extension strategies might be used instead of patch-wise even sampling to utilize the additional channel budget. 
While our focus was on small models supported by the MAX78000/MAX78002 platforms, evaluating larger models could be valuable, given rapid AI hardware advancements.
Regarding societal impact, leveraging additional processors and memory to improve accuracy might increase carbon emissions~\cite{green_ai}, highlighting the need to balance accuracy improvements with environmental sustainability.

 \newpage
\bibliographystyle{plain}
\bibliography{ref}

\begin{thebibliography}{10}

\bibitem{bakar2022protean}
Abu Bakar, Rishabh Goel, Jasper de~Winkel, Jason Huang, Saad Ahmed, Bashima Islam, Przemys{\l}aw Pawe{\l}czak, Kas{\i}m~Sinan Y{\i}ld{\i}r{\i}m, and Josiah Hester.
\newblock Protean: An energy-efficient and heterogeneous platform for adaptive and hardware-accelerated battery-free computing.
\newblock In {\em Proceedings of the 20th ACM Conference on Embedded Networked Sensor Systems}, pages 207--221, 2022.

\bibitem{food101}
Lukas Bossard, Matthieu Guillaumin, and Luc Van~Gool.
\newblock Food-101 -- mining discriminative components with random forests.
\newblock In {\em European Conference on Computer Vision}, 2014.

\bibitem{bottou2010large}
L{\'e}on Bottou.
\newblock Large-scale machine learning with stochastic gradient descent.
\newblock In {\em Proceedings of COMPSTAT'2010: 19th International Conference on Computational StatisticsParis France, August 22-27, 2010 Keynote, Invited and Contributed Papers}, pages 177--186. Springer, 2010.

\bibitem{cai2019once}
Han Cai, Chuang Gan, Tianzhe Wang, Zhekai Zhang, and Song Han.
\newblock Once-for-all: Train one network and specialize it for efficient deployment.
\newblock In {\em International Conference on Learning Representations}, 2020.

\bibitem{cai2018proxylessnas}
Han Cai, Ligeng Zhu, and Song Han.
\newblock Proxylessnas: Direct neural architecture search on target task and hardware.
\newblock In {\em International Conference on Learning Representations}, 2019.

\bibitem{caronti2023fine}
Luca Caronti, Khakim Akhunov, Matteo Nardello, Kas{\i}m~Sinan Y{\i}ld{\i}r{\i}m, and Davide Brunelli.
\newblock Fine-grained hardware acceleration for efficient batteryless intermittent inference on the edge.
\newblock {\em ACM Transactions on Embedded Computing Systems}, 22(5):1--19, 2023.

\bibitem{choi2018pact}
Jungwook Choi, Zhuo Wang, Swagath Venkataramani, Pierce I-Jen Chuang, Vijayalakshmi Srinivasan, and Kailash Gopalakrishnan.
\newblock Pact: Parameterized clipping activation for quantized neural networks.
\newblock {\em arXiv preprint arXiv:1805.06085}, 2018.

\bibitem{coralMicro}
{Google Coral Micro}.
\newblock \url{https://coral.ai/products/dev-board-micro/}. Accessed: 20 May. 2024.

\bibitem{imagenet}
Jia Deng, Wei Dong, Richard Socher, Li-Jia Li, Kai Li, and Li~Fei-Fei.
\newblock Imagenet: A large-scale hierarchical image database.
\newblock In {\em 2009 IEEE Conference on Computer Vision and Pattern Recognition}, pages 248--255, 2009.

\bibitem{fedorov2019sparse}
Igor Fedorov, Ryan~P Adams, Matthew Mattina, and Paul Whatmough.
\newblock Sparse: Sparse architecture search for cnns on resource-constrained microcontrollers.
\newblock {\em Advances in Neural Information Processing Systems}, 32, 2019.

\bibitem{caltech101}
Li~Fei-Fei, Rob Fergus, and Pietro Perona.
\newblock Learning generative visual models from few training examples: An incremental bayesian approach tested on 101 object categories.
\newblock {\em Computer Vision and Pattern Recognition Workshop}, 2004.

\bibitem{gap89}
{Greenwaves Technology}.
\newblock \url{https://greenwaves-technologies.com/low-power-processor/}. Accessed: 20 May. 2024.

\bibitem{gong2023collaborative}
Taesik Gong, Si~Young Jang, Utku~G{\"u}nay Acer, Fahim Kawsar, and Chulhong Min.
\newblock Collaborative inference via dynamic composition of tiny ai accelerators on mcus.
\newblock {\em arXiv preprint arXiv:2401.08637}, 2023.

\bibitem{caltech256}
Gregory Griffin, Alex Holub, and Pietro Perona.
\newblock Caltech-256 object category dataset.
\newblock 2007.

\bibitem{han2015deep_compression}
Song Han, Huizi Mao, and William~J Dally.
\newblock Deep compression: Compressing deep neural networks with pruning, trained quantization and huffman coding.
\newblock {\em International Conference on Learning Representations (ICLR)}, 2016.

\bibitem{hasanpour2016lets}
Seyyed~Hossein Hasanpour, Mohammad Rouhani, Mohsen Fayyaz, and Mohammad Sabokrou.
\newblock Lets keep it simple, using simple architectures to outperform deeper and more complex architectures.
\newblock {\em arXiv preprint arXiv:1608.06037}, 2016.

\bibitem{he2017channel}
Yihui He, Xiangyu Zhang, and Jian Sun.
\newblock Channel pruning for accelerating very deep neural networks.
\newblock In {\em Proceedings of the IEEE international conference on computer vision}, pages 1389--1397, 2017.

\bibitem{imagenette}
Jeremy Howard.
\newblock Imagenette.
\newblock \url{https://github.com/fastai/imagenette/}. Accessed: 20 May. 2024.

\bibitem{analogdevicesinc_ai8x_synthesis}
Analog~Devices Inc.
\newblock Ai8x synthesis repository.
\newblock \url{https://github.com/analogdevicesinc/ai8x-synthesis}, 2024.
\newblock Accessed: 20 May. 2024.

\bibitem{analogdevicesinc_ai8x_training}
Analog~Devices Inc.
\newblock Ai8x training repository.
\newblock \url{https://github.com/analogdevicesinc/ai8x-training}, 2024.
\newblock Accessed: 20 May. 2024.

\bibitem{jacob2018quantization}
Benoit Jacob, Skirmantas Kligys, Bo~Chen, Menglong Zhu, Matthew Tang, Andrew Howard, Hartwig Adam, and Dmitry Kalenichenko.
\newblock Quantization and training of neural networks for efficient integer-arithmetic-only inference.
\newblock In {\em Proceedings of the IEEE conference on computer vision and pattern recognition}, pages 2704--2713, 2018.

\bibitem{adam}
Diederik Kingma and Jimmy Ba.
\newblock Adam: A method for stochastic optimization.
\newblock In {\em International Conference on Learning Representations (ICLR)}, San Diega, CA, USA, 2015.

\bibitem{kwon2023tinytrain}
Young~D Kwon, Rui Li, Stylianos~I Venieris, Jagmohan Chauhan, Nicholas~D Lane, and Cecilia Mascolo.
\newblock Tinytrain: Deep neural network training at the extreme edge.
\newblock {\em arXiv preprint arXiv:2307.09988}, 2023.

\bibitem{liberis2021munas}
Edgar Liberis, {\L}ukasz Dudziak, and Nicholas~D Lane.
\newblock $\mu$nas: Constrained neural architecture search for microcontrollers.
\newblock In {\em Proceedings of the 1st Workshop on Machine Learning and Systems}, pages 70--79, 2021.

\bibitem{liberis2023differentiable}
Edgar Liberis and Nicholas~D Lane.
\newblock Differentiable neural network pruning to enable smart applications on microcontrollers.
\newblock {\em Proceedings of the ACM on Interactive, Mobile, Wearable and Ubiquitous Technologies}, 6(4):1--19, 2023.

\bibitem{lin2021mcunetv2}
Ji~Lin, Wei-Ming Chen, Han Cai, Chuang Gan, and Song Han.
\newblock Mcunetv2: Memory-efficient patch-based inference for tiny deep learning.
\newblock {\em arXiv preprint arXiv:2110.15352}, 2021.

\bibitem{lin2017runtime}
Ji~Lin, Yongming Rao, Jiwen Lu, and Jie Zhou.
\newblock Runtime neural pruning.
\newblock {\em Advances in neural information processing systems}, 30, 2017.

\bibitem{lin2022device}
Ji~Lin, Ligeng Zhu, Wei-Ming Chen, Wei-Chen Wang, Chuang Gan, and Song Han.
\newblock On-device training under 256kb memory.
\newblock {\em Advances in Neural Information Processing Systems}, 35:22941--22954, 2022.

\bibitem{liu2018intriguing}
Rosanne Liu, Joel Lehman, Piero Molino, Felipe Petroski~Such, Eric Frank, Alex Sergeev, and Jason Yosinski.
\newblock An intriguing failing of convolutional neural networks and the coordconv solution.
\newblock {\em Advances in neural information processing systems}, 31, 2018.

\bibitem{liu2017multi}
Shuo Liu and Zheng Liu.
\newblock Multi-channel cnn-based object detection for enhanced situation awareness.
\newblock {\em arXiv preprint arXiv:1712.00075}, 2017.

\bibitem{liu2019metapruning}
Zechun Liu, Haoyuan Mu, Xiangyu Zhang, Zichao Guo, Xin Yang, Kwang-Ting Cheng, and Jian Sun.
\newblock Metapruning: Meta learning for automatic neural network channel pruning.
\newblock In {\em Proceedings of the IEEE/CVF international conference on computer vision}, pages 3296--3305, 2019.

\bibitem{liu2017learning}
Zhuang Liu, Jianguo Li, Zhiqiang Shen, Gao Huang, Shoumeng Yan, and Changshui Zhang.
\newblock Learning efficient convolutional networks through network slimming.
\newblock In {\em Proceedings of the IEEE international conference on computer vision}, pages 2736--2744, 2017.

\bibitem{max32650}
{Analog MAX32650}.
\newblock \url{https://www.analog.com/en/products/max32650.html}. Accessed: 20 May. 2024.

\bibitem{max78000}
{Analog MAX78000}.
\newblock \url{https://www.analog.com/en/products/max78000.html}. Accessed: 20 May. 2024.

\bibitem{max78000benchmark}
{Cutting the AI Power Cord: Technology to Enable True Edge Inference}.
\newblock \url{https://cms.tinyml.org/wp-content/uploads/talks2020/tinyML_Talks_Kris_Ardis_and_Robert_Muchsel_-201027.pdf}. Accessed: 20 May. 2024.

\bibitem{max78000fth}
{Analog MAX78000FTHR}.
\newblock \url{https://www.analog.com/en/design-center/evaluation-hardware-and-software/evaluation-boards-kits/max78000fthr.html}. Accessed: 20 May. 2024.

\bibitem{max78002}
{Analog MAX78002}.
\newblock \url{https://www.analog.com/en/products/max78002.html}. Accessed: 20 May. 2024.

\bibitem{max78002evkit}
{Analog MAX78002EVKIT}.
\newblock \url{https://www.analog.com/en/design-center/evaluation-hardware-and-software/evaluation-boards-kits/max78002evkit.html}. Accessed: 20 May. 2024.

\bibitem{moosmann2023tinyissimoyolo}
Julian Moosmann, Marco Giordano, Christian Vogt, and Michele Magno.
\newblock Tinyissimoyolo: A quantized, low-memory footprint, tinyml object detection network for low power microcontrollers.
\newblock In {\em 2023 IEEE 5th International Conference on Artificial Intelligence Circuits and Systems (AICAS)}, pages 1--5. IEEE, 2023.

\bibitem{moss2022ultra}
Arthur Moss, Hyunjong Lee, Lei Xun, Chulhong Min, Fahim Kawsar, and Alessandro Montanari.
\newblock Ultra-low power dnn accelerators for iot: Resource characterization of the max78000.
\newblock In {\em Proceedings of the 20th ACM Conference on Embedded Networked Sensor Systems}, pages 934--940, 2022.

\bibitem{paszke2019pytorch}
Adam Paszke, Sam Gross, Francisco Massa, Adam Lerer, James Bradbury, Gregory Chanan, Trevor Killeen, Zeming Lin, Natalia Gimelshein, Luca Antiga, et~al.
\newblock Pytorch: An imperative style, high-performance deep learning library.
\newblock {\em Advances in neural information processing systems}, 32, 2019.

\bibitem{rastegari2016xnor}
Mohammad Rastegari, Vicente Ordonez, Joseph Redmon, and Ali Farhadi.
\newblock Xnor-net: Imagenet classification using binary convolutional neural networks.
\newblock In {\em European conference on computer vision}, pages 525--542. Springer, 2016.

\bibitem{ruegg2023kp2dtiny}
Thomas R{\"u}egg, Marco Giordano, and Michele Magno.
\newblock Kp2dtiny: Quantized neural keypoint detection and description on the edge.
\newblock In {\em 2023 IEEE 5th International Conference on Artificial Intelligence Circuits and Systems (AICAS)}, pages 1--5. IEEE, 2023.

\bibitem{rusci2020memory}
Manuele Rusci, Alessandro Capotondi, and Luca Benini.
\newblock Memory-driven mixed low precision quantization for enabling deep network inference on microcontrollers.
\newblock {\em Proceedings of Machine Learning and Systems}, 2:326--335, 2020.

\bibitem{sandler2018mobilenetv2}
Mark Sandler, Andrew Howard, Menglong Zhu, Andrey Zhmoginov, and Liang-Chieh Chen.
\newblock Mobilenetv2: Inverted residuals and linear bottlenecks.
\newblock In {\em Proceedings of the IEEE conference on computer vision and pattern recognition}, pages 4510--4520, 2018.

\bibitem{green_ai}
Roy Schwartz, Jesse Dodge, Noah~A. Smith, and Oren Etzioni.
\newblock Green ai.
\newblock {\em Commun. ACM}, 63(12):54–63, nov 2020.

\bibitem{stm32f7}
{STM32F7 Series}.
\newblock \url{https://www.st.com/en/microcontrollers-microprocessors/stm32f7-series.html}. Accessed: 20 May. 2024.

\bibitem{tan2021efficientnetv2}
Mingxing Tan and Quoc Le.
\newblock Efficientnetv2: Smaller models and faster training.
\newblock In {\em International conference on machine learning}, pages 10096--10106. PMLR, 2021.

\bibitem{wang2019haq}
Kuan Wang, Zhijian Liu, Yujun Lin, Ji~Lin, and Song Han.
\newblock Haq: Hardware-aware automated quantization with mixed precision.
\newblock In {\em Proceedings of the IEEE/CVF conference on computer vision and pattern recognition}, pages 8612--8620, 2019.

\bibitem{wang2018depth}
Weiyue Wang and Ulrich Neumann.
\newblock Depth-aware cnn for rgb-d segmentation.
\newblock In {\em Proceedings of the European conference on computer vision (ECCV)}, pages 135--150, 2018.

\bibitem{wang2018location}
Zhenyi Wang and Olga Veksler.
\newblock Location augmentation for cnn.
\newblock {\em arXiv preprint arXiv:1807.07044}, 2018.

\bibitem{zheng2024streamnet}
Hong-Sheng Zheng, Yu-Yuan Liu, Chen-Fong Hsu, and Tsung~Tai Yeh.
\newblock Streamnet: Memory-efficient streaming tiny deep learning inference on the microcontroller.
\newblock {\em Advances in Neural Information Processing Systems}, 36, 2024.

\bibitem{zhou2016dorefa}
Shuchang Zhou, Yuxin Wu, Zekun Ni, Xinyu Zhou, He~Wen, and Yuheng Zou.
\newblock Dorefa-net: Training low bitwidth convolutional neural networks with low bitwidth gradients.
\newblock {\em arXiv preprint arXiv:1606.06160}, 2016.

\bibitem{zhu2016trained}
Chenzhuo Zhu, Song Han, Huizi Mao, and William~J Dally.
\newblock Trained ternary quantization.
\newblock In {\em International Conference on Learning Representations}, 2016.

\end{thebibliography}

\clearpage

\appendix

\section{Experimental details}~\label{app:exp_details}


For all experiments conducted in the paper, we used three different random seeds (0, 1, 2) and reported the average accuracy with standard deviations.  

\subsection{Tiny AI accelerator platforms}\label{app:exp_details:accelerator}

\begin{figure}[h]
    \centering
    \begin{subfigure}[t]{0.38\textwidth}
        \centering
        \includegraphics[width=\linewidth]{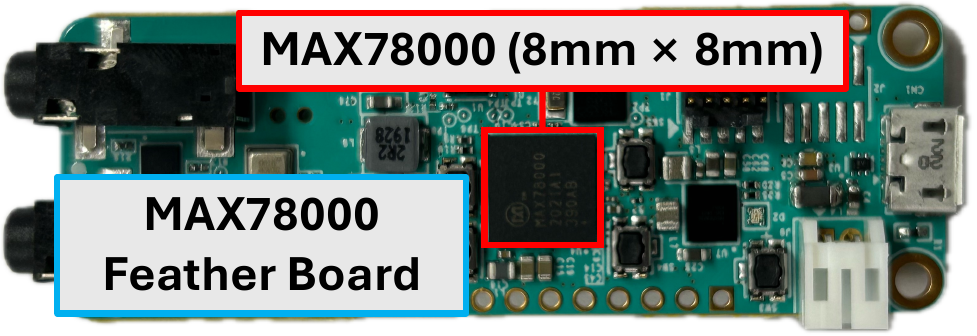}
        \caption{MAX78000 Feather Board~\cite{max78000fth}.}
        \label{fig:max78000}
    \end{subfigure}
    \hspace{0.1cm}
    \begin{subfigure}[t]{0.6\textwidth}
        \centering
        \includegraphics[width=\linewidth]{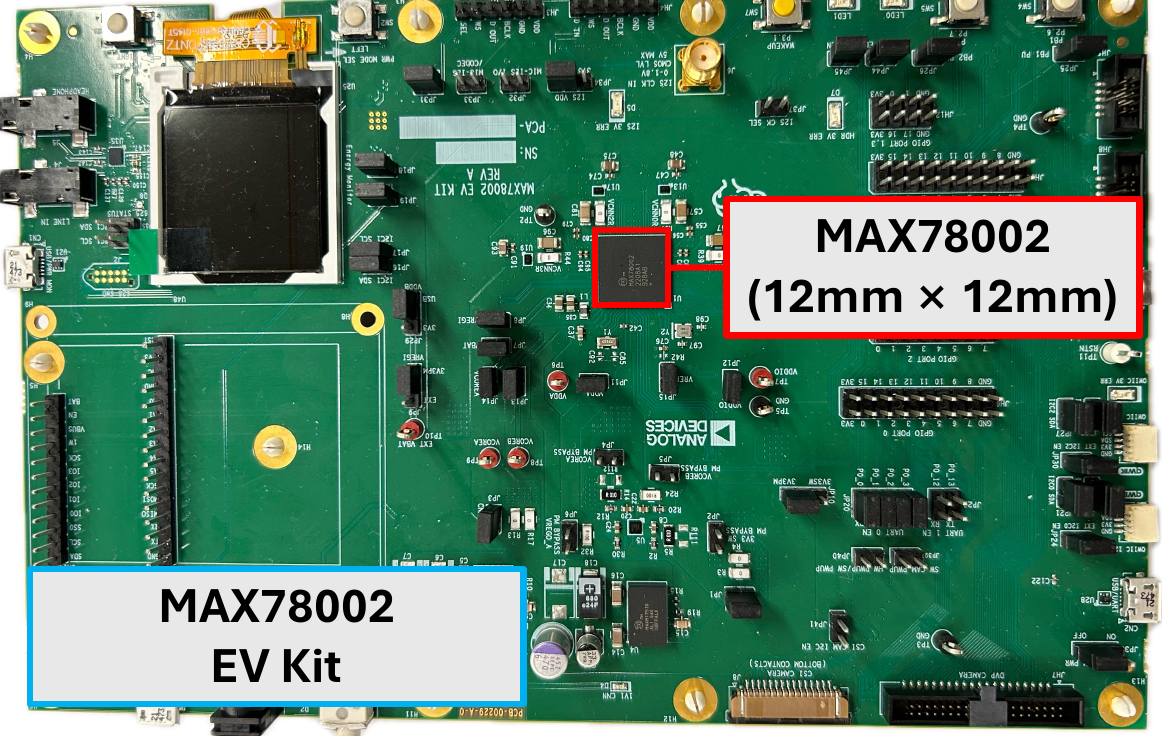}
        \caption{MAX78002 Evaluation (EV) Kit~\cite{max78002evkit}.}
        \label{fig:max78002}
    \end{subfigure}
    \caption{Two tiny AI accelerator development platforms used in our work. Note that although the development platform is bulky, the actual size of the accelerators is tiny (e.g., 8mm$\times$8mm for MAX78000). All data storing and model inference is done only in the AI Accelerator part (MAX78000 and MAX78002). }\label{fig:testbed}
\end{figure}

\begin{table}[ht]
\centering
\caption{Comparison of MAX78000 and MAX78002.}
\begin{tabular}{ccc}
\toprule
\textbf{Component} & \textbf{MAX78000}~\cite{max78000} & \textbf{MAX78002}~\cite{max78002} \\ \midrule
MCU Processor & Arm Cortex-M4 (100 MHz), RISC-V & Arm Cortex-M4 (120 MHz), RISC-V \\ 
Flash Memory & 512 KB & 2.5 MB \\ 
SRAM & 128 KB & 384 KB \\ 
CNN Processor & 64 parallel CNN processors & 64 parallel CNN processors \\ 
Data Memory & 512 KB & 1.3 MB \\ 
Weight Memory & 432 KB & 2 MB \\ 
Bias Memory & 2 KB & 8 KB \\ 
\bottomrule
\end{tabular}
\label{tab:max_comparison}
\end{table}

In this paper, we focus on the MAX78000~\cite{max78000} and MAX78002~\cite{max78002} as our primary platforms since they are the most popular research platforms~\cite{moosmann2023tinyissimoyolo, ruegg2023kp2dtiny, gong2023collaborative, bakar2022protean, caronti2023fine, moss2022ultra} owing to the disclosed hardware details and open-source tools, enabling in-depth analysis and modification of their operations. Figure~\ref{fig:testbed} shows our testbed. Note that all operations required for the model inference are done under the AI accelerator part (highlighted with the red boxes), while the entire boards are bigger for development purposes. For on-device deployment and measurement, we used MAX78000 for SimpleNet and WideNet and MAX78002~\cite{max78002} for EfficientNetV2 and MobileNetV2, following the memory requirement. For the sake of explanation, we assumed each processor is mapped to one memory instance in this paper, although MAX78000/MAX78002 group four data memory instances together to communicate with four processors in reality. Our experiments were conducted on the actual implementations. Table~\ref{tab:max_comparison} compares MAX78000 and MAX78002.



\subsection{Model training}~\label{app:exp_details:training}

We followed the official model training code for MAX78000 and MAX78002 platforms~\cite{analogdevicesinc_ai8x_training}. Here, we detail the hyperparameters used in the official training code. We trained models with NVIDIA A40 GPUs.

For all models, quantization-aware training is conducted with support for batch normalization after convolutional layers through batch normalization fusing~\cite{jacob2018quantization}. This fusing operation integrates the effects of batch normalization directly into the parameters of the preceding convolutional layer by adjusting the weights and bias values. Consequently, after the fusing/folding process, the network no longer contains any batch normalization layers. Instead, the effects of batch normalization are reflected in the modified weights and biases of the preceding convolutional layers.

\paragraph{SimpleNet.} 
SimpleNet~\cite{hasanpour2016lets} was trained for 300 epochs using the Adam optimizer~\cite{adam} with an initial learning rate of 0.001 and a batch size of 32. A multi-step learning rate scheduler was used with milestones set at epochs 100, 150, and 200, and a multiplicative factor of learning rate decay value of 0.25. Quantization-aware training (QAT) was introduced starting at epoch 240. During QAT, a shift quantile of 0.985 was applied to manage activation ranges. The weight precision was primarily set to 2 bits. However, exceptions were made for certain layers: the 1st convolutional layer utilized 8-bit weights, while the 2nd, 11th, 12th, 13th, and 14th convolutional layers used 4-bit weights.

\paragraph{WideNet.}
WideNet~\cite{hasanpour2016lets} was trained for 300 epochs using the Adam optimizer~\cite{adam} with an initial learning rate of 0.001 and a batch size of 100. A multi-step learning rate scheduler was used with milestones set at epochs 100, 150, and 200, and a multiplicative factor of learning rate decay value of 0.25. Quantization-aware training (QAT) was introduced starting at epoch 240. During QAT, a shift quantile of 0.985 was applied to manage activation ranges. The weight precision was primarily set to 2 bits. However, exceptions were made for certain layers: the 1st convolutional layer utilized 8-bit weights, while the 2nd, 11th, 12th, 13th, and 14th convolutional layers used 4-bit weights.

\paragraph{EfficientNetV2.}
EfficientNetV2~\cite{tan2021efficientnetv2} was trained for 300 epochs using the Adam optimizer~\cite{adam} with an initial learning rate of 0.001 and a batch size of 100. A multi-step learning rate scheduler was used with milestones set at epochs 50, 100, 150, 200, and 250 and a multiplicative factor of learning rate decay value of 0.5.  Quantization-aware training (QAT) was introduced starting at epoch 210. During QAT, a shift quantile of 0.995 was applied to manage activation ranges. The weight precision was primarily set to 8 bits. 

\paragraph{MobileNetV2.}
MobileNetV2~\cite{sandler2018mobilenetv2} was trained for 300 epochs using the stochastic gradient descent optimizer (SGD)~\cite{bottou2010large} with an initial learning rate of 0.1 and a batch size of 128. A multi-step learning rate scheduler was used with milestones set at epochs 100, 150, 175, and 250 and a multiplicative factor of learning rate decay value of 0.235. Quantization-aware training (QAT) was introduced starting at epoch 200. During QAT, a shift quantile of 1.0 was applied to manage activation ranges. The weight precision was primarily set to 8 bits.

\subsection{Datasets}

\paragraph{ImageNette.} Imagenette~\cite{imagenette} is a smaller, more manageable subset of ImageNet~\cite{imagenet}, containing 10 classes. These classes include tench, English springer, cassette player, chain saw, church, French horn, garbage truck, gas pump, golf ball, and parachute. ImageNette has 9469/3925 train/test samples with the original image shape of $3\times350\times350$. All images were normalized with the ImageNet mean (0.485, 0.456, 0.406) and standard deviations (0.229, 0.224, 0.225), and then converted to Q7 format (one byte per data) to support on-device inference with the tiny AI accelerator platforms (MAX78000 and MAX78002).

\paragraph{Caltech101.} Caltech101~\cite{caltech101} is a dataset composed of images representing objects from 101 different categories, in addition to a background clutter category. Each image features a single object and is labeled accordingly. The number of images per category ranges from approximately 40 to 800, resulting in a total of around 8677 images. 
Caltech101 has 6941/1736 train/test samples and the original image shape of $3\times300\times300$. All images were normalized with the ImageNet mean (0.485, 0.456, 0.406) and standard deviations (0.229, 0.224, 0.225), and then converted to Q7 format (one byte per data) to support on-device inference with the tiny AI accelerator platforms (MAX78000 and MAX78002).

\paragraph{Caltech256.} Caltech256~\cite{caltech256} built upon its previous version, Caltech101, offering enhancements such as larger category sizes, additional and more extensive clutter categories, and increased overall difficulty. 
The dataset contains 29780 images across 256 classes after removing the clutter class. 
Caltech256 has 23824/5956 train/test samples with the original image shape of $3\times300\times300$. All images were normalized with the ImageNet mean (0.485, 0.456, 0.406) and standard deviations (0.229, 0.224, 0.225), and then converted to Q7 format (one byte per data) to support on-device inference with the tiny AI accelerator platforms (MAX78000 and MAX78002).

\paragraph{Food101.} Food101~\cite{food101} includes 101 food categories, each with 750 images for training and 250 images for testing, which is a total of 101000 images. The original images were rescaled to have a maximum side length of 512 pixels. This dataset has 75750/25250 train/test samples and the original image with the shape of $3\times512\times512$. All images were normalized with the ImageNet mean (0.485, 0.456, 0.406) and standard deviations (0.229, 0.224, 0.225), and then converted to Q7 format (one byte per data) to support on-device inference with the tiny AI accelerator platforms (MAX78000 and MAX78002).

\subsection{Baseline details}\label{app:baseline:details}

\paragraph{Downsampling.} Downsampling is a straightforward method that collects samples evenly distributed across the original image. This approach is equivalent to the case when the number of channels is equal to three in \system{}.

\paragraph{CoordConv.} CoordConv~\cite{liu2018intriguing} pointed out the limitation of traditional CNNs that relied solely on RGB images for the coordinate transformation problem and introduced the augmentation of $i$ and $j$ coordinates, which improved object detection efficiency. We referred to the Pytorch implementation of CoordConv\footnote{\url{https://github.com/walsvid/CoordConv}} for implementing this baseline. 

\paragraph{CoordConv (with r).} The authors of CoordConv also introduced the third channel for an $r$ coordinate, where $r = \sqrt{(i - h/2)^2 + (j - w/2)^2}$, which they found effective in some experiments. Similar to CoordConv, we referred to the Pytorch implementation of CoordConv for implementing this baseline.

\subsection{Alternative data channel extension methods' details }\label{app:alternative:details}

\begin{figure}[h]
    \centering
    \includegraphics[width=1\linewidth]{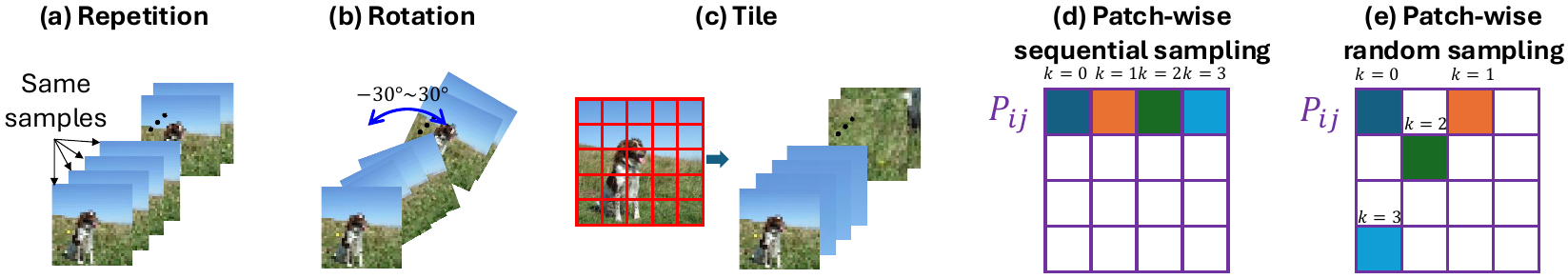}
    \caption{Visulaization of four alternative data extension methods.}
    \label{fig:alternative}
\end{figure}

\paragraph{Repetition.} Repetition (Figure~\ref{fig:alternative}(a)) repeats the same downsampled image across the channels until it reaches the maximum possible number of input channels, which is the same as the number of data memory instances (64). 

\paragraph{Rotation.} Rotation (Figure~\ref{fig:alternative}(b)) generates slightly different images through rotating images from the downsampled image. It makes rotated images until it reaches the maximum possible number of input channels. The angle of rotation ranges from -30 to 30 degrees. For instance, given a downsampled three-channel image input and target channel size of 64, it generates rotated images with an angle linearly spaced between -30 to 30 degrees.

\paragraph{Tile.} Tile (Figure~\ref{fig:alternative}(c)) divides the original image into multiple tiles and stacks those tiles across channels. Specifically, given the number of images take $K = \lceil \frac{C_O}{C_I} \rceil $, it finds the nearest square number $S$ that is higher than or equal to $K$, (e.g., $S=5^2 > 22$ when $K=22$). The original image is then divided into equal-sized patches. Each patch is subsequently downsampled to the target size. The downsampled patches are collected and concatenated along the channel dimension, forming a new image with the desired number of channels. If the total number of patches exceeds the target number of channels, the excess patches are discarded. 

\paragraph{Patch-wise sequential sampling.} Patch-wise sequential sampling (Figure~\ref{fig:alternative}(d)) is similar to \system{} but it involves sequential sampling within a patch instead of even sampling. Specifically, it samples the first $K$ samples for each patch and follows the same channel-wise stacking procedure in \system{}.

\paragraph{Patch-wise random sampling.} Patch-wise random sampling (Figure~\ref{fig:alternative}(e)) is similar to \system{} but it involves random sampling within a patch instead of even sampling. Specifically, it samples randomly-selected $K$ samples for each patch and follows the same channel-wise stacking procedure in \system{}.

\section{Additional Experiments}\label{app:experiment}

\subsection{Overhead of the channel expansion on devices}

\paragraph{Channel expansion latency.}
The latency of the channel expansion process depends on the processor's computational capability. During our evaluation, we pre-processed data on a powerful server, and thus data processing was negligible. We additionally conducted data processing on the ultra-low-power MCU processor on the board (Arm Cortex-M4) to understand the data processing overhead on less-capable devices. We measured the overhead of applying DEX to expand channels from a $3\times224\times224$ image (a typical size for ImageNet) to $64\times32\times32$ (the highest channel expansion used in our accelerators) on the MAX78002's Arm Cortex-M4 (120MHz).

This process took 2.2 ms on the Arm Cortex-M4. In terms of memory, this addition took the SRAM memory of 62KB ($64\times32\times32$ Bytes - $3\times32\times32$ Bytes) on the processor. However, since \system{} extends data to a size that the data memory in the AI accelerator can accommodate, this additional memory will not be an issue from the AI accelerator’s perspective.

\paragraph{Impact on end-to-end inference performance.} Note that the MCU processor and the AI accelerator are independent processing components that run in parallel. This means that if the inference latency on the accelerator is higher than the data processing latency, data can be pre-processed for the next inference during the current inference (and thus data processing latency can be hidden). For inference, the inference latency of EfficientNet (11.7ms) is higher than the data processing latency of 2.2ms, and thus the inference throughput remains the same under continuous inference.

However, this depends on the scenario. The end-to-end impact of data processing latency depends on the processor's computational capability, the dimension of the data, and the size of channel expansion. For instance, in scenarios where data processing is done and transferred in more capable machines (e.g., cloud servers, smartphones, etc.) than the MCU processor on the tiny AI accelerator, the impact of data processing can be even more negligible.

\subsection{Power consumption}

We measured the power consumption of the inference on MAX78000 by varying the size of the channel extension with a Monsoon Power Monitor. The result is shown in Table~\ref{tab:power}. 
As the number of channels increased, power consumption increased accordingly. This is because a higher number of channels uses more processors in the AI accelerator, leading to increased power consumption. 

\begin{table}[h]
\centering
\caption{The power consumption of inference measured by varying the size of the channel extension with a Monsoon Power Monitor. All numbers are in milliwatts (mW).}
\begin{tabular}{lrrrrr}
\toprule
\textbf{Model} & \multicolumn{1}{c}{\textbf{Chan = 3}} & \multicolumn{1}{c}{\textbf{Chan = 6}} & \multicolumn{1}{c}{\textbf{Chan = 18}} & \multicolumn{1}{c}{\textbf{Chan = 36}} & \multicolumn{1}{c}{\textbf{Chan = 64}} \\
\midrule
SimpleNet & 53.82 & 53.85 & 58.21 & 61.42 & 68.9 \\
WideNet & 60.74 & 61.37 & 63.76 & 67.92 & 77.14\\
\bottomrule
\end{tabular}%
\label{tab:power}
\end{table}

\section{Example images generated from \system{}} 

\begin{figure}[h]
    \centering
    \includegraphics[width=1\linewidth]{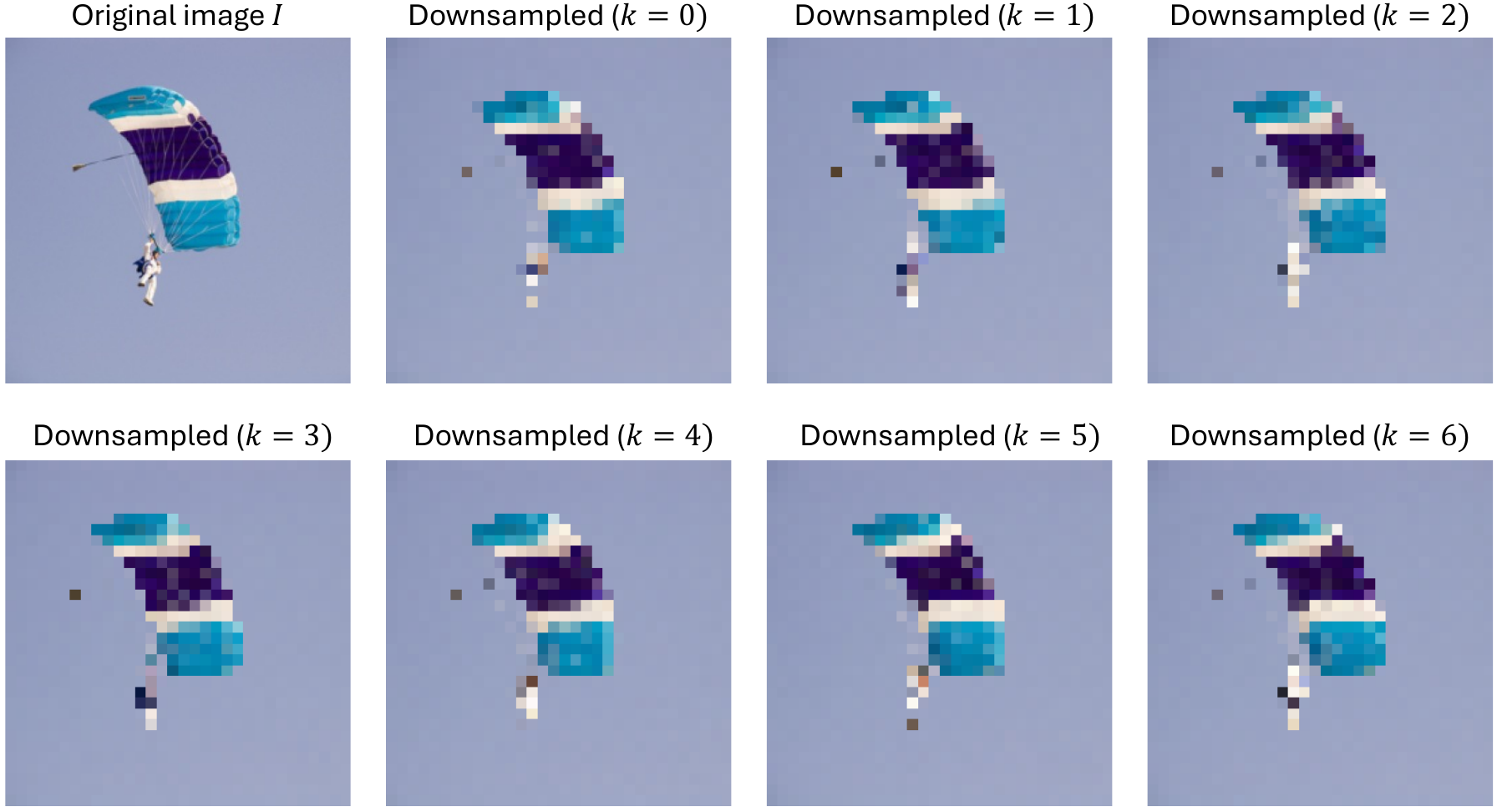}
    \caption{Examples images generated from an original $3\times350\times350$ image (ImageNette) to $3\times32\times32$ downsampled image via \system{}. $k=0$ to $k=6$ cases are shown only. Each generated image contains different pixel information, which collectively enhances feature learning in CNNs.}
    \label{fig:example1}
\end{figure}

\begin{figure}[h]
    \centering
    \includegraphics[width=1\linewidth]{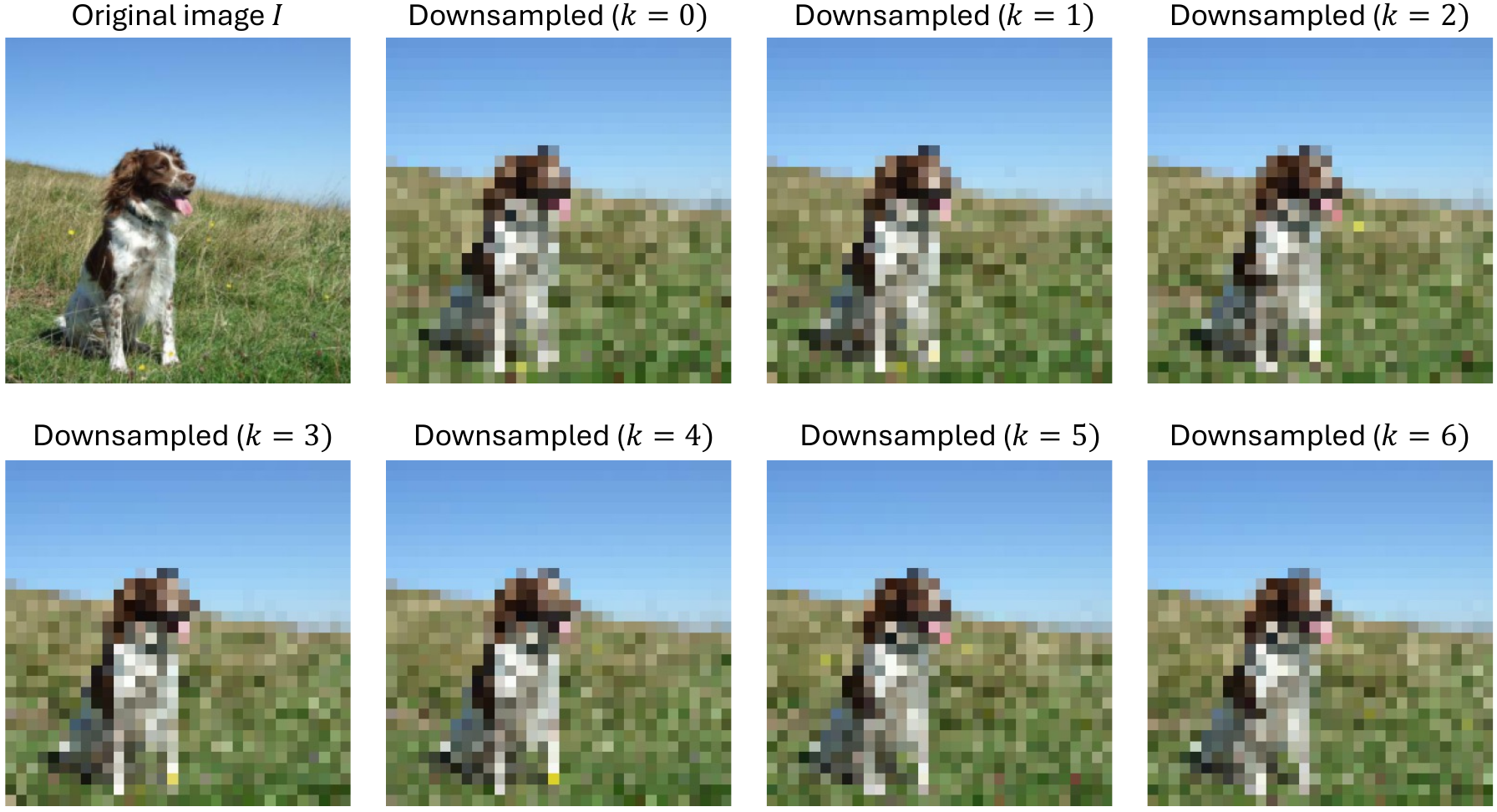}
    \caption{Examples images generated from an original $3\times350\times350$ image (ImageNette) to $3\times32\times32$ downsampled image via \system{}. $k=0$ to $k=6$ cases are shown only. Each generated image contains different pixel information, which collectively enhances feature learning in CNNs.}
    \label{fig:example2}
\end{figure}

\begin{figure}[h]
    \centering
    \includegraphics[width=1\linewidth]{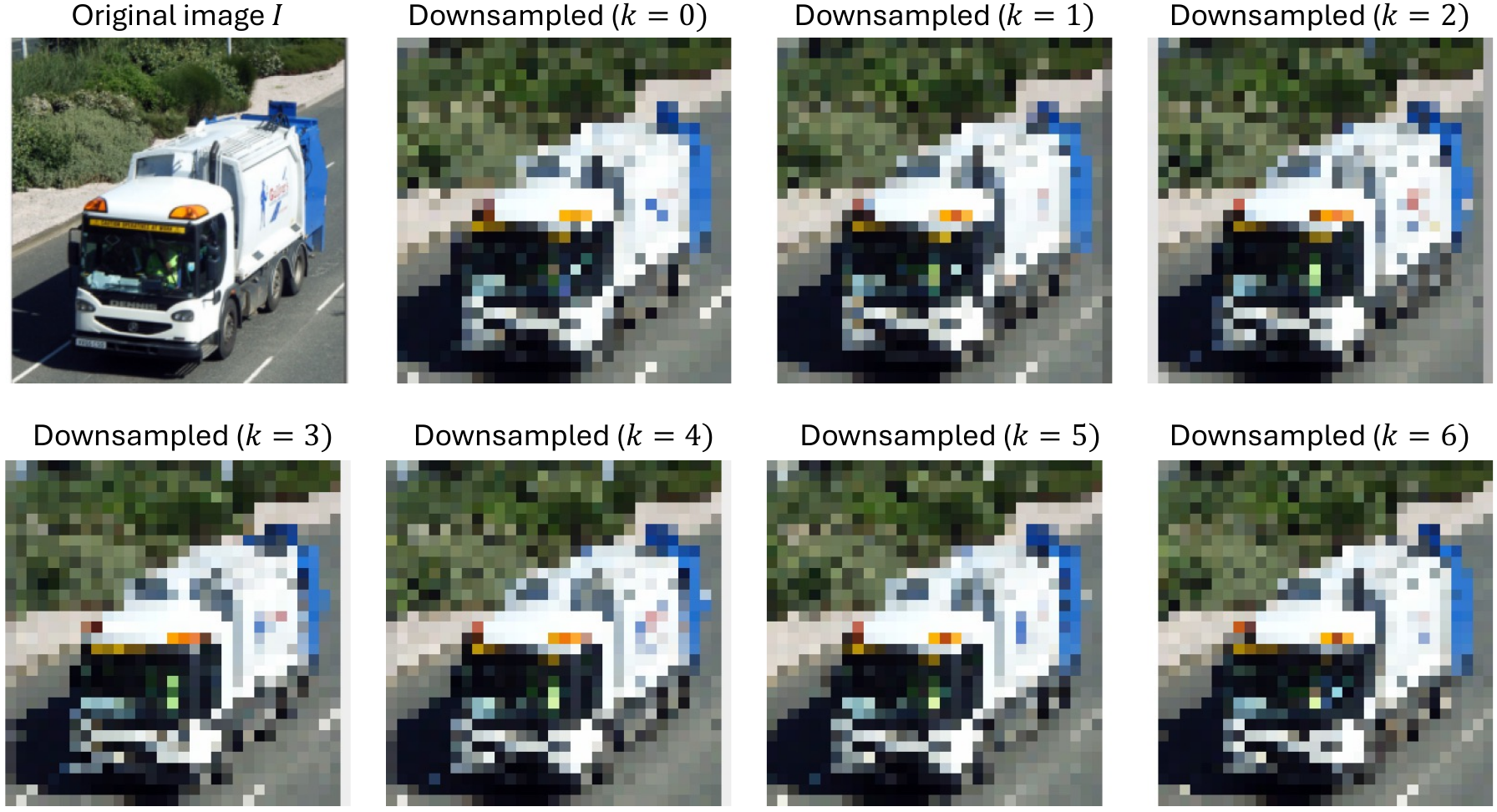}
    \caption{Examples images generated from an original $3\times350\times350$ image (ImageNette) to $3\times32\times32$ downsampled image via \system{}. $k=0$ to $k=6$ cases are shown only. Each generated image contains different pixel information, which collectively enhances feature learning in CNNs.}
    \label{fig:example3}
\end{figure}





\section{License of assets}\label{app:license}

\paragraph{Datasets.} ImageNette dataset (Apache-2.0 license), Caltech101 (CC BY 4.0), Caltech256 (CC BY 4.0), and Food101 dataset (MIT license).

\paragraph{Codes.} 
AI8X-training for MAX78000 and MAX78002 (Apache-2.0 license), AI8X-synthesis for MAX78000 and MAX78002 (Apache-2.0 license), and PyTorch implementation of CoordConv (MIT license).

\clearpage

\newpage

\section*{NeurIPS Paper Checklist}

The checklist is designed to encourage best practices for responsible machine learning research, addressing issues of reproducibility, transparency, research ethics, and societal impact. Do not remove the checklist: {\bf The papers not including the checklist will be desk rejected.} The checklist should follow the references and precede the (optional) supplemental material.  The checklist does NOT count towards the page
limit. 

Please read the checklist guidelines carefully for information on how to answer these questions. For each question in the checklist:
\begin{itemize}
    \item You should answer \answerYes{}, \answerNo{}, or \answerNA{}.
    \item \answerNA{} means either that the question is Not Applicable for that particular paper or the relevant information is Not Available.
    \item Please provide a short (1–2 sentence) justification right after your answer (even for NA). 
\end{itemize}

{\bf The checklist answers are an integral part of your paper submission.} They are visible to the reviewers, area chairs, senior area chairs, and ethics reviewers. You will be asked to also include it (after eventual revisions) with the final version of your paper, and its final version will be published with the paper.

The reviewers of your paper will be asked to use the checklist as one of the factors in their evaluation. While "\answerYes{}" is generally preferable to "\answerNo{}", it is perfectly acceptable to answer "\answerNo{}" provided a proper justification is given (e.g., "error bars are not reported because it would be too computationally expensive" or "we were unable to find the license for the dataset we used"). In general, answering "\answerNo{}" or "\answerNA{}" is not grounds for rejection. While the questions are phrased in a binary way, we acknowledge that the true answer is often more nuanced, so please just use your best judgment and write a justification to elaborate. All supporting evidence can appear either in the main paper or the supplemental material, provided in appendix. If you answer \answerYes{} to a question, in the justification please point to the section(s) where related material for the question can be found.

IMPORTANT, please:
\begin{itemize}
    \item {\bf Delete this instruction block, but keep the section heading ``NeurIPS paper checklist"},
    \item  {\bf Keep the checklist subsection headings, questions/answers and guidelines below.}
    \item {\bf Do not modify the questions and only use the provided macros for your answers}.
\end{itemize}


\begin{enumerate}

\item {\bf Claims}
    \item[] Question: Do the main claims made in the abstract and introduction accurately reflect the paper's contributions and scope?
    \item[] Answer: \answerYes{} 
    \item[] Justification: The main claims in the abstract and introduction accurately reflect the paper's contributions and scope.

\item {\bf Limitations}
    \item[] Question: Does the paper discuss the limitations of the work performed by the authors?
    \item[] Answer: \answerYes{} 
    \item[] Justification: See \S\ref{sec:discussion}.

\item {\bf Theory Assumptions and Proofs}
    \item[] Question: For each theoretical result, does the paper provide the full set of assumptions and a complete (and correct) proof?
    \item[] Answer: \answerNA{} 
    \item[] Justification: No theoretical result.

    \item {\bf Experimental Result Reproducibility}
    \item[] Question: Does the paper fully disclose all the information needed to reproduce the main experimental results of the paper to the extent that it affects the main claims and/or conclusions of the paper (regardless of whether the code and data are provided or not)?
    \item[] Answer: \answerYes{} 
    \item[] Justification: Experimental details are in \S\ref{sec:evaluation} and Appendix~\ref{app:exp_details}.

\item {\bf Open access to data and code}
    \item[] Question: Does the paper provide open access to the data and code, with sufficient instructions to faithfully reproduce the main experimental results, as described in supplemental material?
    \item[] Answer: \answerYes{} 
    \item[] Justification: Yes, the source code is available at \url{https://github.com/Nokia-Bell-Labs/data-channel-extension}.

\item {\bf Experimental Setting/Details}
    \item[] Question: Does the paper specify all the training and test details (e.g., data splits, hyperparameters, how they were chosen, type of optimizer, etc.) necessary to understand the results?
    \item[] Answer: \answerYes{} 
    \item[] Justification: Experimental details are in \S\ref{sec:evaluation} and Appendix~\ref{app:exp_details}.

\item {\bf Experiment Statistical Significance}
    \item[] Question: Does the paper report error bars suitably and correctly defined or other appropriate information about the statistical significance of the experiments?
    \item[] Answer: \answerYes{} 
    \item[] Justification: See~\ref{sec:eval:result}. We ran the experiments with three random seems (0,1,2) and reported the standard deviations.

\item {\bf Experiments Compute Resources}
    \item[] Question: For each experiment, does the paper provide sufficient information on the computer resources (type of compute workers, memory, time of execution) needed to reproduce the experiments?
    \item[] Answer:  \answerYes{} 
    \item[] Justification: See \S\ref{sec:background}, \S\ref{sec:evaluation}, and Appendix~\ref{app:exp_details}.
    
\item {\bf Code Of Ethics}
    \item[] Question: Does the research conducted in the paper conform, in every respect, with the NeurIPS Code of Ethics \url{https://neurips.cc/public/EthicsGuidelines}?
    \item[] Answer: \answerYes{} 
    \item[] Justification: We follow the NeurIPS Code of Ethics.

\item {\bf Broader Impacts}
    \item[] Question: Does the paper discuss both potential positive societal impacts and negative societal impacts of the work performed?
    \item[] Answer: \answerYes{} 
    \item[] Justification: See \S\ref{sec:discussion}.
    
\item {\bf Safeguards}
    \item[] Question: Does the paper describe safeguards that have been put in place for responsible release of data or models that have a high risk for misuse (e.g., pretrained language models, image generators, or scraped datasets)?
    \item[] Answer: \answerNA{} 
    \item[] Justification: No such components.

\item {\bf Licenses for existing assets}
    \item[] Question: Are the creators or original owners of assets (e.g., code, data, models), used in the paper, properly credited and are the license and terms of use explicitly mentioned and properly respected?
    \item[] Answer: \answerYes{} 
    \item[] Justification: See \S\ref{sec:evaluation} and Appendix~\ref{app:license}.

\item {\bf New Assets}
    \item[] Question: Are new assets introduced in the paper well documented and is the documentation provided alongside the assets?
    \item[] Answer: \answerNA{} 
    \item[] Justification: No new assets.

\item {\bf Crowdsourcing and Research with Human Subjects}
    \item[] Question: For crowdsourcing experiments and research with human subjects, does the paper include the full text of instructions given to participants and screenshots, if applicable, as well as details about compensation (if any)? 
    \item[] Answer: \answerNA{} 
    \item[] Justification: No crowdsourcing conducted. 

\item {\bf Institutional Review Board (IRB) Approvals or Equivalent for Research with Human Subjects}
    \item[] Question: Does the paper describe potential risks incurred by study participants, whether such risks were disclosed to the subjects, and whether Institutional Review Board (IRB) approvals (or an equivalent approval/review based on the requirements of your country or institution) were obtained?
    \item[] Answer: \answerNA{} 
    \item[] Justification: No IRB required.

\end{enumerate}

\end{document}